\documentclass[a4paper,11pt]{article}
\usepackage[top=4cm, bottom=4cm, left=3cm, right=3cm]{geometry}

\usepackage[utf8]{inputenc} 
\usepackage[T1]{fontenc}    
\usepackage{hyperref}       
\usepackage{url}            
\usepackage{booktabs}       
\usepackage{amsfonts}       
\usepackage{nicefrac}       
\usepackage{microtype}      

\usepackage{wrapfig}

\usepackage{subcaption}
\usepackage{caption}

\usepackage{amsthm}
\usepackage{mathtools}
\usepackage{amsmath}
\usepackage{amssymb}
\usepackage{bm}
\usepackage{upgreek}

\usepackage{algorithm}
\usepackage{algorithmic}

\usepackage{setspace}
\usepackage{xspace}

\usepackage{enumitem}
\usepackage{siunitx}

\usepackage{cleveref}

\usepackage{todonotes}
\definecolor{blued}{RGB}{70,197,221}

\definecolor{citrine}{rgb}{0.89, 0.82, 0.04}

\newcommand{\Nystrom}[1]{{Nystr\"om}}

\providecommand{\scal}[2]{\left\langle{#1},{#2}\right\rangle}
\providecommand{\nor}[1]{\lVert{#1}\rVert}

\providecommand{\tr}{\operatorname{Tr}}

\newcommand{\R}{\mathbb R}

\newcommand{\N}{\mathbb N}

\newcommand{\hh}{\mathcal H}

\newcommand{\la}{\lambda}

\newcommand{\expect}[1]{{\mathbb E}[#1]}

\newcommand{\X}{{X}}
\newcommand{\Y}{{Y}}

\newcommand{\rhox}{{\rho_{\X}}}

\newcommand{\C}{{C}}

\newcommand{\Cn}{\widehat{\C}_n}

\newcommand{\Sn}{\widehat{S}_n}

\newcommand{\Km}{K_{MM}}
\newcommand{\Knm}{K_{nM}}
\newcommand{\yn}{\widehat{y}}

\newcommand{\fh}{f_\hh}

\newcommand{\M}{M}

\newcommand{\emplev}{\widehat{\ell}}

\newcommand{\alglev}{\widetilde{\ell}}

\newcommand{\eqals}[1]{\begin{align*}#1\end{align*}}
\newcommand{\eqal}[1]{\begin{align}#1\end{align}}
\newcommand{\bpr}{\begin{proof}}
	\newcommand{\epr}{\end{proof}}
\newcommand{\be}{\begin{equation}}
\newcommand{\ee}{\end{equation}}

\newtheorem{definition}{Definition}
\newcommand{\bd}{\begin{definition}}
	\newcommand{\ed}{\end{definition}}

\newcommand{\bi}{\begin{itemize}}
	\newcommand{\ei}{\end{itemize}}

\newtheorem{ass}{Assumption}
\newcommand{\ba}{\begin{ass}}
	\newcommand{\ea}{\end{ass}}

\newtheorem{remark}{Remark}
\newcommand{\br}{\begin{remark}}
	\newcommand{\er}{\end{remark}}

\newtheorem{example}{Example}
\newcommand{\bex}{\begin{example}}
	\newcommand{\eex}{\end{example}}

\newtheorem{proposition}{Proposition}
\newcommand{\bp}{\begin{proposition}}
	\newcommand{\ep}{\end{proposition}}

\newtheorem{lemma}{Lemma}
\newcommand{\blm}{\begin{lemma}}
	\newcommand{\elm}{\end{lemma}}

\newtheorem{theorem}{Theorem}
\newcommand{\bt}{\begin{theorem}}
	\newcommand{\et}{\end{theorem}}

\newtheorem{corollary}{Corollary}
\newcommand{\bcor}{\begin{corollary}}
	\newcommand{\ecor}{\end{corollary}}

\date{}

\usepackage[symbol]{footmisc}
\renewcommand{\thefootnote}{\fnsymbol{footnote}}

\def\:#1{\protect \ifmmode {\mathbf{#1}} \else {\textbf{#1}} \fi}

\newcommand{\wt}[1]{\widetilde{#1}}
\newcommand{\wh}[1]{\widehat{#1}}

\newcommand{\nystrom}{Nystr\"{o}m\xspace}

\renewcommand{\algorithmicrequire}{\textbf{Input:}}
\renewcommand{\algorithmicensure}{\textbf{Output:}}

\newcounter{cnt-lem-quad-variation}
\newcommand{\whp}{w.h.p.\@\xspace}

\newcommand{\bigotime}{\mathcal{O}}

\newcommand{\probability}{\mathbb{P}}

\DeclareMathOperator*{\expectedvalue}{\mathbb{E}}

\newcommand{\uniformdist}{Uniform}

\newcommand{\multinomialdist}{Multinomial}

\newcommand{\deff}{{d_{\text{eff}}}}

\title{On Fast Leverage Score Sampling and Optimal Learning}

\author{Alessandro Rudi $^{*, 1}$~~~\\ 
{\small\em alessandro.rudi@inria.fr~~~~~~}
\and ~~~~~~Daniele Calandriello $^{*, 2}$\\ 
{\small\em ~~daniele.calandriello@iit.it} 
\and Luigi Carratino $^{3}$~~~~~~~~\\
{\small\em luigi.carratino@dibris.unige.it~~~~~~~~~~}
\and ~~Lorenzo Rosasco $^{2,3}$~~~~~~\\
{\small\em ~~lrosasco@mit.edu}~~~~~~~~~}

\begin{document}

\footnotetext[1]{Equal contribution.}
\renewcommand{\thefootnote}{\arabic{footnote}}
\footnotetext[1]{INRIA -- Sierra-project team \& École Normale Supérieure, Paris.}
\footnotetext[2]{LCSL -- Istituto Italiano di Tecnologia, Genova, Italy \& MIT, Cambridge, USA.}
\footnotetext[3]{DIBRIS -- Universit\`a degli Studi di Genova, Genova, Italy.}
\maketitle

\begin{abstract}
 Leverage score sampling provides an appealing way to perform approximate computations for large matrices.
Indeed, it allows to derive  faithful approximations with a  complexity  adapted to the  problem at hand. 
Yet, performing leverage scores  sampling  is a challenge in its own right requiring further approximations.
In this paper, we study the problem of leverage score sampling for positive definite matrices defined by a kernel. 
Our contribution is twofold. First we provide a novel   algorithm for  leverage score sampling and 
second, we exploit the proposed method in  statistical learning by 
deriving a novel solver for kernel ridge regression. Our main technical contribution is showing that the proposed algorithms 
are currently the most efficient and accurate for these problems.
\end{abstract}

\section{Introduction}
A variety of machine learning problems require manipulating and performing computations with large matrices that often do not fit memory.
In practice, randomized techniques are often employed to reduce the computational burden. Examples include stochastic approximations \cite{arora2012stochastic}, columns/rows subsampling  and more general sketching techniques \cite{woodruff_sketching_2014,tropp2012user}. 
One of  the simplest approach is uniform column sampling \cite{williams2001using,bach2013sharp},  that is  replacing the original matrix with a subset of columns chosen uniformly at random. This approach   is fast to compute, but the number  of columns needed  for a prescribed approximation accuracy does not take advantage  of the possible low rank structure of the  matrix at hand. As discussed in \cite{alaoui2015fast},  leverage score sampling  provides a way to tackle this shortcoming.
Here columns are sampled proportionally  to suitable weights, called leverage scores (LS) \cite{drineas_fast_2012,alaoui2015fast}.  With this sampling strategy,  the number of columns needed for a prescribed accuracy is governed by the so called {\em effective dimension} which is a natural extension of the notion of rank. Despite these nice properties, performing leverage score sampling provides a challenge in its own right, since it has complexity in the same order of an eigendecomposition of the original matrix. Indeed, much effort has been recently devoted to derive fast and provably accurate algorithms for  approximate leverage score sampling  \cite{woodruff_sketching_2014,calandriello_disqueak_2017,alaoui2015fast,musco2016provably, pmlr-v70-calandriello17a}. 
\\
In this paper, we consider these questions in the case of positive semi-definite matrices, central for example
in Gaussian processes \cite{rasmussen_gaussian_2006} and kernel methods \cite{scholkopf2002learning}. 
Sampling approaches in this context are related to the so called \nystrom approximation \cite{smola2000sparse} and \nystrom centers selection problem \cite{rasmussen_gaussian_2006}, and are widely studied both in practice \cite{williams2001using} and in theory \cite{bach2013sharp}.
Our contribution is twofold. First, we propose and study BLESS, a novel algorithm for approximate leverage scores sampling.
The first solution to this problem is introduced in \cite{alaoui2015fast}, but has poor approximation guarantees and high time  complexity. Improved approximations are achieved by algorithms recently proposed in \cite{calandriello_disqueak_2017} and \cite{musco2016provably}.
In particular, the approach in \cite{calandriello_disqueak_2017} can obtain good accuracy and  very efficient computations but only as long as distributed resources are available. Our first  technical contribution is showing  that our algorithm can achieve state of the art accuracy and computational complexity without requiring distributed resources.  The key  idea is to follow a coarse to fine strategy, alternating uniform and leverage scores sampling on sets of increasing size.  \\
Our second, contribution is considering leverage score sampling in statistical learning  with  least squares.
We extend  the  approach  in \cite{rudi2017falkon} for efficient kernel ridge regression based on   combining
fast optimization algorithms (preconditioned conjugate gradient) with uniform sampling. Results in \cite{rudi2017falkon} showed that optimal learning bounds can be achieved with a complexity which is $\wt{\bigotime}(n\sqrt{n})$ in time  and $\wt{\bigotime}(n)$ space. In this paper, we study  the impact of replacing uniform with leverage score sampling.  In particular, we prove that the derived method  still achieves optimal learning bounds but the
time and memory is now  $\wt{\bigotime}(n \deff)$, and $\wt{\bigotime}(\deff^2)$ respectively, where $\deff$ is the effective dimension which and is never larger, and possibly much smaller, than $\sqrt{n}$. To the best of our knowledge this is the best currently known computational guarantees for a kernel ridge regression solver.  \section{Leverage score sampling with BLESS}
After introducing leverage score sampling  and previous algorithms, we present our approach and first theoretical results. 
\subsection{Leverage score sampling}
Suppose $\widehat{K} \in \R^{n\times n}$ is symmetric and  positive semidefinite. A basic   question is 
deriving memory efficient approximation of $\widehat K$ \cite{williams2001using,calandriello_disqueak_2017} or related quantities, e.g.
approximate projections on its range \cite{musco2016provably}, or associated estimators, as in  kernel ridge regression \cite{rudi2015less,rudi2017falkon}. The eigendecomposition of $\widehat{K}$ offers a natural, but  computationally demanding solution.
Subsampling columns  (or rows) is  an appealing alternative.  A  basic approach is uniform sampling, whereas a more refined approach is leverage scores sampling. This latter procedure corresponds to 
sampling  columns with probabilities proportional  to the leverage scores 
\be\label{eq:lev_scores}
\ell(i,\lambda) = \left(\wh{K}(\wh{K} + \lambda n I)^{-1}\right)_{ii},\quad \quad \quad i\in [n],
\ee
where  $[n]=\{1, \dots, n\}$.
The advantage of leverage score sampling, is that potentially very few columns can suffice for the desired   approximation. 
Indeed, letting  
$$
d_\infty (\lambda) = n\max_{i=1, \dots, n} \ell(i,\lambda), \quad \quad \quad 
\deff(\lambda) = \sum_{i=1}^n\ell(i,\lambda), 
$$
 for $\la>0$, it is easy to see that $\deff(\lambda) \le d_\infty (\lambda)\le 1/\la$
 for all $\la$, and  previous results show that the number of columns required for 
 accurate approximation
are $d_\infty$  for uniform sampling and  $\deff$ for leverage score sampling \cite{bach2013sharp,alaoui2015fast}. 
However,  it is clear from definition~\eqref{eq:lev_scores} that an exact  leverage scores computation
 would require the same order of computations as an eigendecomposition, hence
 approximations are needed.The accuracy of approximate leverage scores is typically
 measured by $t>0$ in multiplicative bounds of the form
\eqal{\label{eq:cond-appr-lev}
\frac{1}{1+t} \ell(i,\la) \leq \alglev(i,\la) \leq (1+t)\ell(i,\la), \quad \forall i \in [n].
}
Before proposing a new improved solution, we briefly discuss relevant previous
works.  To provide a unified view, some preliminary discussion is useful. 
 
\subsection{Approximate leverage scores}
First, we recall how a subset of columns can be used to compute approximate  leverage scores. 
For $M\le n$,  let $J = \{j_i\}_{i=1}^M$ with $j_i \in [n] $, and  $\wh{K}_{J,J} \in \mathbb{R}^{M \times M}$ with entries $(K_{J,J})_{lm} = K_{{j_l},{j_m}}$. 
For $i\in [n] $, let $\wh{K}_{J,i} = (\wh{K}_{{j_1},i},\dots, \wh{K}_{j_M,i})$ and 
consider for $\la>1/n$,
\be\label{def:out-of-sample-lev-scores}
\alglev_{J}(i,\lambda) =(\la n)^{-1} (\wh{K}_{ii} -  \wh{K}_{J,i}^\top (\wh{K}_{J,J} + \lambda n A)^{-1} \wh{K}_{J,i}),
\ee
where $A\in \mathbb{R}^{M \times M}$ is a matrix to be specified
\footnote{Clearly, $\alglev_{J}$ depends on the choice of the matrix $A$, but we omit this dependence  to simplify the notation.} (see  later for details). The above definition is motivated by the observation that if
$J= [n]$, and $A=I$,  then $\alglev_{J}(i,\lambda)= \ell(i,\la)$, by the following identity
$$
\widehat{K}(\widehat{K} + \lambda n I)^{-1} = (\la n)^{-1}(\widehat{K} - \widehat{K}(\widehat{K} + \lambda n I)^{-1}\widehat{K}).
$$
In the following, it is also useful to  consider a subset of leverage scores computed as in~\eqref{def:out-of-sample-lev-scores}.
For $M\le R\le n$,  let $U =  \{u_i\}_{i=1}^R$ with $u_i \in [n] $, and  
\be\label{eq:metaLSS}
L_J(U, \la)= \{\alglev_{J}(u_1,\lambda), \dots, \alglev_{J}(u_R,\lambda) \}.
\ee
Also in the following we will use the notation 
\be\label{metaLSS}
L_J(U, \la) \mapsto J'
\ee
to indicate the leverage score sampling of   $J'\subset U$ columns based on the leverage scores   $L_J(U, \la)$, that is the procedure of sampling columns from $U$ according to their leverage scores~\ref{eq:lev_scores}, computed using $J$, to obtain a new subset of columns $J'$.
\\
We end noting that leverage score sampling~\eqref{metaLSS} requires  $\bigotime(M^2)$  memory  to store $K_J$, and $\bigotime(M^3+RM^2)$ time 
to invert $K_J$, and  compute $R$ leverage scores via~\eqref{def:out-of-sample-lev-scores}.

\subsection{Previous algorithms for leverage scores computations}
We discuss relevant previous approaches  using the above quantities.
\\
\\
{\bf \textsc{Two-Pass} sampling \cite{alaoui2015fast}}.
This is the  first approximate leverage score sampling  proposed, and is  based on 
 using directly~\eqref{metaLSS} as $L_{J_1}(U_2,\lambda) \mapsto J_2$, with $U_2=  [n]$ and $J_1$ a subset taken uniformly at random. Here we call this method \textsc{Two-Pass} sampling since it requires two rounds of sampling on the whole set $[n]$, one uniform to select $J_1$ and one using leverage scores to select $J_2$.
\\
\\
{\bf \textsc{Recursive-RLS}  \cite{musco2016provably}}. This is a development of \textsc{Two-Pass} sampling based on the idea of recursing the above construction.
In our notation, let  $U_1\subset U_2 \subset U_3= [n]$, where $U_1,U_2$ are uniformly sampled and have cardinalities $n/4$ and $n/2$, respectively. 
The idea is to start from $J_1=U_1$, and consider first
$$
L_{J_1}(U_2, \la) \mapsto J_2,
$$
but then continue with
$$
L_{J_2}(U_3, \la) \mapsto J_3.
$$
Indeed, the above construction can be made recursive for a family of nested subsets $(U_h)_H$ of cardinalities $n/2^{h}$, considering  $J_1=U_1$ and 
\be\label{eq:coniugi}
L_{J_{h}}(U_{h+1}, \la) \mapsto J_{h+1}.
\ee
\\
{\bf \textsc{SQUEAK}\cite{calandriello_disqueak_2017}}. This approach follows a different iterative strategy.
Consider  a partition  $U_1, U_2 , U_3$ of $[n]$, so that  $U_j=n/3$, for $j=1, \dots 3$.
Then,  consider  $J_1= U_1$, and 
  $$
L_{J_1 \cup U_2}(J_1\cup U_2, \la) \mapsto J_2,
$$
and then continue with
$$
L_{J_2 \cup U_3}(J_2 \cup U_3, \la) \mapsto J_3.
$$
Similarly to the other cases, the procedure is iterated considering $H$ subsets $(U_h)_{h=1}^H$ each with cardinality $n/H$. Starting from  $J_1= U_1$ the iterations is
\be\label{eq:squeak}
L_{J_{h} \cup U_{h+1}}(J_{h} \cup U_{h+1}, \la).
\ee

We note that  all the above procedures require specifying  the number of iteration to be performed,  the weights matrix to compute the leverage scores at each iteration, and a strategy to select the subsets $(U_h)_h$. In all the above cases the selection of $U_h$ is based on  uniform sampling, while the number of iterations and weight choices arise from theoretical considerations (see \cite{alaoui2015fast,calandriello_disqueak_2017,musco2016provably} for details).

Note that \textsc{Two-Pass sampling} uses a set $J_1$ of cardinality roughly $1/\la$ (an upper bound on $d_\infty(\la)$) and incurs in a computational cost of $RM^2 = n/\la^2$. In comparison, \textsc{Recursive-RLS} \cite{musco2016provably} leads to essentially the same accuracy while improving computations. In particular, the sets $J_h$ are never larger than $\deff(\la)$. Taking into account that at the last iteration performs leverage score sampling on $U_h = [n]$, the total computational complexity is $n\deff(\la)^2$.
\textsc{SQUEAK} \cite{calandriello_disqueak_2017} recovers the same accuracy, size of $J_h$, and $n\deff(\la)^2$ time complexity when $|U_h| \simeq \deff(\la)$, but only requires a single pass over the data.
We also note that a distributed version of \textsc{SQUEAK} is discussed in \cite{calandriello_disqueak_2017}, which allows to reduce the computational cost to $n \deff(\la)^2/p$, provided $p$ machines are available.

\subsection{Leverage score sampling with BLESS}

The procedure we propose, dubbed BLESS,  has similarities  to the one proposed in \cite{musco2016provably} (see ~\eqref{eq:coniugi}), but also some important differences.
The main difference is that, rather than a fixed $\lambda$, we consider a decreasing sequence of parameters $\la_0 > \la_1 > \dots > \la_H=\la$ resulting in different algorithmic choices.
For the construction of the subsets $U_h$ we do not use nested subsets,
but rather each $(U_h)_{h=1}^H$ is sampled uniformly and independently,
with a size smoothly increasing as $1/\la_h$.
Similarly, as in \cite{musco2016provably} we proceed iteratively, but at each iteration a
different decreasing parameter $\lambda_h$ is used to compute the leverage scores.
Using the notation introduced above, the iteration of BLESS is given by
\be\label{eq:metabless}
L_{J_{h}}(U_{h+1}, \la_{h+1}) \mapsto J_{h+1},
\ee
where the initial set $J_1=U_1$ is sampled uniformly with size roughly $1/\lambda_0$.
\\
BLESS has two main advantages. The first is computational: each of the sets
$U_h$, including the final $U_H$,
has cardinality smaller than $1/\lambda$. Therefore the overall runtime
has a cost of only $RM^2 \leq M^2/\la$, which can be
dramatically smaller than the $nM^2$ cost achieved by the methods
in~\cite{musco2016provably},~\cite{calandriello_disqueak_2017} 
and is comparable to the distributed version of SQUEAK using  $p= \la/n$
machines. The second advantage is that a whole {\em path} of leverage scores $\{\ell(i,\lambda_h)\}_{h=1}^H$
is computed at once, in the sense that at each iteration accurate approximate
leverage scores at scale $\la_h$ are computed. This is extremely useful in practice,
as it can be used when cross-validating $\la_h$. As a comparison, for all
previous method a full run of the algorithm is needed for each  value of $\la_h$.
\begin{algorithm}[t]
	\begin{algorithmic}[1]
		\renewcommand\algorithmiccomment[1]{			\(\triangleright\){#1}}
		\renewcommand{\algorithmicrequire}{\textbf{Input:}}
		\renewcommand{\algorithmicensure}{\textbf{Output:}}
		\REQUIRE dataset $\{x_i\}_{i=1}^n$, regularization $\la$, step $q$, starting reg. $\la_0$, constants $q_1, q_2$ controlling the approximation level.
		\ENSURE $M_h \in [n]$ number of selected points, $J_h$ set of indexes, $A_h$ weights.
		\STATE $J_0 = \emptyset,~A_0 = [],~H = \frac{\log(\la_0/\la)}{\log q}$ \label{line:initialize-optls}
		\FOR{$h = 1 \dots H$}
		\STATE $\la_h = \la_{h-1}/q$
        \STATE set constant $R_h = q_1 \min\{\kappa^2/\la_h, \;n\}$
		\STATE sample $U_h = \{u_1, \dots, u_{R_h}\}$ i.i.d. $u_i \sim \uniformdist([n])$
        \STATE compute $\alglev_{J_{h-1}}(x_{u_k}, \la_h)$ for all $u_k \in U_h$ using Eq.~\ref{def:out-of-sample-lev-scores}
		\STATE set $P_h = (p_{h,k})_{k=1}^{R_h}$ with $p_{h,k} = \alglev_{J_{h-1}}(x_{u_k}, \la_h)/(\sum_{u \in U_h} \alglev_{J_{h-1}}(x_u, \la_h))$
		\STATE set constant $M_h =  q_2 d_h$ with $d_h = \frac{n}{R_h}\sum_{u \in U_h} \alglev_{J_{h-1}}(x_u, \la_h)$, and 
		\STATE sample $J_h = \{j_1,\dots, j_{M_h}\}$ i.i.d. $j_i \sim \multinomialdist(P_h, U_h)$
		\STATE $A_h = \frac{R_h M_h}{n}\textrm{diag}\left(p_{h,j_1}, \dots, p_{h, j_{M_h}}\right)$
		\ENDFOR
	\end{algorithmic}
	\caption{Bottom-up Leverage Scores Sampling (BLESS)
	}
	\label{alg:est-ridge-lev-scores}
\end{algorithm}
\begin{algorithm}[t]
	\begin{algorithmic}[1]
		\renewcommand\algorithmiccomment[1]{			\(\triangleright\){#1}}
		\renewcommand{\algorithmicrequire}{\textbf{Input:}}
		\renewcommand{\algorithmicensure}{\textbf{Output:}}
		\REQUIRE dataset $\{x_i\}_{i=1}^n$, regularization $\la$, step $q$, starting reg. $\la_0$, constant $q_2$ controlling the approximation level.
		\ENSURE $M_h \in [n]$ number of selected points, $J_h$ set of indexes, $A_h$ weights.
        \STATE 
		$J_0 = \emptyset, ~A_0 = [], ~ H = \frac{\log(\la_0/\la)}{\log q}$,
		\FOR{$h = 1 \dots H$}
		\STATE $\la_h = \la_{h-1}/q$
        \STATE set constant $\beta_h = \min\{q_2\kappa^2/(\la_h n), \;1\}$
        \STATE initialize $U_h = \emptyset$
        \FOR{$i \in [n]$}
            \STATE add $i$ to $U_h$ with probability $\beta_h$
        \ENDFOR
        \FOR{$j \in U_h$}
        \STATE compute $p_{h,j} = \min\{q_2\alglev_{J_{h-1}}(x_{j}, \la_{h-1}), \; 1\}$
        \STATE add $j$ to $J_h$ with probability $p_{h,j}/\beta_h$\label{line:rls_rejection}
        \ENDFOR
		\STATE $J_h = \{j_1,\dots, j_{M_h}\}$, and $A_h = \textrm{diag}\left(p_{h,j_1}, \dots, p_{h, j_{M_h}}\right).$
		\ENDFOR
	\end{algorithmic}
\caption{Bottom-up  Leverage Scores Sampling without Replacement (BLESS-R)}
\label{alg:mahoneyplus-nystrom-fast}
\end{algorithm}
\\
In the paper we consider two variations of the above general idea leading to
Algorithm~\ref{alg:est-ridge-lev-scores} and
Algorithm~\ref{alg:mahoneyplus-nystrom-fast}.  The main difference in the two
algorithms lies in the way in which sampling is performed:  with and without
replacement, respectively. In particular, considering sampling without
replacement (see~\ref{alg:mahoneyplus-nystrom-fast}) it is possible to take the
set $(U_h)_{h=1}^H$ to be nested and also to obtain slightly improved results, as shown in the next section.
\\
The derivation of  BLESS rests on some basic ideas. First,  note that, since sampling uniformly a set 
$U_\la$ of size $d_\infty(\la)\le 1/\la$ allows a good approximation, then  we can 
replace $L_{[n]}([n], \la)\mapsto J$ by 
\be\label{2pass+}
L_{U_\la}(U_\la, \la)\mapsto J,
\ee
 where $J$ can be taken to have cardinality $\deff(\la)$.
 However,  this is still costly, and the idea is to  repeat  and  couple 
approximations at multiple scales. Consider $\la'>\la$, a set $U_{\la'}$ of size $d_\infty(\la')\le  1/\la'$ sampled uniformly,  and 
$L_{U_{\la'}}(U_{\la'}, \la')\mapsto J'$. 
The basic idea behind BLESS is to replace~\eqref{2pass+} by
$$
L_{J'}(U_\la, \la)\mapsto \tilde J.
$$
The key result, see ,  is that  taking $\tilde J$ of cardinality 
\be\label{magni}
(\la'/\la) \deff(\la)
\ee 
suffice to achieve the same accuracy as $J$. 
Now,  if we take $\la'$ sufficiently large,  it is easy to see that
$\deff(\la')\sim d_\infty(\la')\sim 1/\la'$, so that we can take $J'$ uniformly at random. However, the factor $(\la'/\la)$ in~\eqref{magni} becomes too big.
Taking multiple scales fix this problem and leads to the iteration in~\eqref{eq:metabless}. \subsection{Theoretical guarantees}\label{sect:worst}
Our first  main result establishes  in a precise and quantitative way the advantages of BLESS.

\bt\label{thm:main-appr-lev-scores}
Let $n \in \N$, $\la > 0$ and $\delta \in (0,1]$. Given $t > 0, q > 1$ and $H \in \N$, $(\la_h)_{h = 1}^H$ defined as in \Cref{alg:est-ridge-lev-scores,alg:mahoneyplus-nystrom-fast}, when $(J_h, a_h)_{h=1}^H$ are computed
\begin{enumerate}
	\item by Alg.~\ref{alg:est-ridge-lev-scores} with parameters $\la_0 = \frac{\kappa^2}{\min(t,1)}$,~ $q_1 \geq\frac{5 \kappa^2 q_2}{q(1+t)}$,~ $q_2 \geq 12 q {(2t + 1)^2 \over t^2}  (1+t)\log \frac{12 H n}{\delta}$,
	\item by Alg.~\ref{alg:mahoneyplus-nystrom-fast} with parameters $\la_0 = \frac{\kappa^2}{\min(t,1)}$,~ $q_1 \geq 54 \kappa^2 {(2t + 1)^2 \over t^2} \log \frac{12 H n}{\delta}$,
\end{enumerate}
let $\alglev_{J_h}(i, \la_h)$ as in Eq.~\eqref{def:out-of-sample-lev-scores} depending on $J_h, A_h$, then with probability at least $1-\delta$:
\begin{itemize}
	\item[(a)] $\displaystyle\qquad \frac{1}{1+t} \ell(i, \la_h) ~~\leq~~ \alglev_{J_h}(i, \la_h) ~~\leq~~ (1 + \min(t,1))\ell(i, \la_h), \quad \forall i \in [n], h \in [H],$
		\item[(b)] $\displaystyle\qquad  |J_h| \leq q_2 \deff(\la_h),  \quad \forall h \in [H].$
\end{itemize}
\et

The  above result confirms that the subsets $J_h$ computed by BLESS are accurate in the desired sense, see~\eqref{eq:cond-appr-lev}, and the size of all $J_h$ is small and proportional to $\deff(\lambda_h)$,  leading to a  computational cost of only $\bigotime\left(\min\left(\frac{1}{\la},n\right)\deff(\la)^2\log^2\frac{1}{\la}\right)$ in  time and $O\left(\deff(\la)^2\log^2\frac{1}{\la}\right)$ in space (for additional properties of $J_h$ see Thm.~\ref{thm:alg-appr-lev-scores-extended-form} in appendixes). \Cref{table:comparison-appr-lev} compares the complexity and number of columns sampled by BLESS with other methods. The crucial point is that in most applications, the parameter $\la$ is chosen as a decreasing function of $n$, e.g. $\la= 1/\sqrt{n}$, resulting in potentially massive computational gains. 
Indeed, since BLESS computes leverage scores for sets of size at most $1/\la$, this allows to perform leverage scores sampling on matrices with millions of rows/columns, as shown in the experiments. In the next section, we illustrate the impact of BLESS in the context of supervised statistical learning. 
\begin{table}
	\renewcommand{\arraystretch}{1.2}
	\centering
	\begin{tabular}{@{}l c l@{}}
		\toprule
		Algorithm & Runtime & $|J|$\\
		\midrule
				Uniform Sampling \cite{bach2013sharp} & $-$ & $1/\la$ \\
				Exact RLS Sampl.\@ & $n^3$ & $\deff(\lambda)$\\
				Two-Pass Sampling \cite{alaoui2015fast} & $n/\lambda^2$ & $\deff(\la)$  \\
				Recursive RLS  \cite{musco2016provably}& $n\deff(\lambda)^2$ & $\deff(\lambda)$ \\
				SQUEAK \cite{calandriello_disqueak_2017}& $n\deff(\lambda)^2$ & $\deff(\lambda)$\\
				This work, Alg. \ref{alg:est-ridge-lev-scores} and \ref{alg:mahoneyplus-nystrom-fast} & $\bm{1/\lambda ~d_{\rm eff}(\lambda)^2}$ & $\deff(\lambda)$\\
		\bottomrule
	\end{tabular}
	\caption{\footnotesize The proposed algorithms are compared with the state of the art (in $\wt{\bigotime}$ notation), in terms of time complexity and cardinality of the set $J$ required to satisfy the approximation condition in Eq.~\ref{eq:cond-appr-lev}. \label{table:comparison-appr-lev}}
\end{table} \section{Efficient supervised learning with leverage scores}\label{sect:stat}
In this section, we discuss the impact of BLESS in a supervised  learning. Unlike most previous results on leverage scores sampling in this context \cite{alaoui2015fast,calandriello_disqueak_2017,musco2016provably},  we consider  the   setting of statistical learning, where the challenge is that inputs, as well as the outputs, are random.
More precisely, given a probability space $(\X\times \Y, \rho)$,
where $\Y\subset \R$, and  considering least squares, the problem is to  solve
\eqal{\label{eq:learning-problem}
	\min_{f \in \hh} {\cal E}(f),\quad {\cal E}(f) = \int_{\X\times\Y} (f(x)-y)^2 d\rho(x,y),
}
when $\rho$ is known only through $(x_i,y_i)_{i=1}^n\sim \rho^n$.
In the above minimization problem, $\hh$ is a reproducing kernel Hilbert space defined by a 
positive definite kernel $K:\X\times \X\to \R$ \cite{scholkopf2002learning}. Recall that the latter is defined as the  completion of $\text{span}\{K(x, \cdot) ~|~x\in \X\}$  with  the inner product  $\scal{K(x,\cdot)}{K(x',\cdot)}_\hh = K(x,x')$. 
The quality of an empirical approximate solution $\widehat{f}$ is  measured via probabilistic bounds on the {excess risk} ${\cal R}(\widehat{f}~) = {\cal E}(\widehat{f}~) - \min_{f \in \hh}{\cal E}(f).$
\subsection{Learning with FALKON-BLESS}
The algorithm we propose,  called FALKON-BLESS, combines BLESS with FALKON  \cite{rudi2017falkon} a state of the art  algorithm  to solve the least squares problem presented above. The appeal of FALKON is that it is currently the most efficient solution to achieve optimal excess risk bounds. As we discuss in the following,  the combination with BLESS leads to further improvements. \\ 
We describe the derivation of the considered algorithm  starting from  kernel ridge regression  (KRR) 
\be\label{krr}
\widehat f_\la (x)= \sum_{i=1}^n K(x,x_i)c_i , \quad \quad c= (\widehat{K}+\la n I)^{-1}\widehat{Y} 
\ee
where $c= (c_1, \dots, c_n)$, $\widehat{Y} = (y_1, \dots, y_n)$ and $\widehat{K}\in \R^{n\times n }$ is the empirical kernel matrix with entries $(\widehat{K})_{ij}=K(x_i, x_j)$. KRR has optimal statistical properties \cite{caponnetto}, but large $\bigotime(n^3)$ time and $\bigotime(n^2)$ space requirements.
FALKON can be seen as an  approximate ridge regression solver combining a number of algorithmic ideas.
First, sampling is used to select a subset  $\{\widetilde{x}_1, \dots, \widetilde{x}_M \}$ of the input  data uniformly at random,
and to define  an approximate solution 
\be\label{nkrr}
\widehat{f}_{\la,M}(x) = \sum_{j=1}^M  K(\widetilde{x}_j, x)\alpha_j, \quad \quad \alpha =(K_{nM}^\top K_{nM} + \la K_{MM})^{-1} K_{nM}^\top y,
\ee
where $\alpha=(\alpha_1, \dots, \alpha_M)$,  $K_{nM} \in \R^{n\times M}$, has entries $(K_{nM})_{ij} = K(x_i ,\tilde{x}_j)$ and $K_{MM} \in \R^{M\times \M}$ has entries $(K_{MM})_{jj'} = K(\tilde{x}_j, \tilde{x}_{j'})$, with $i \in [n], j,j' \in [M]$. We note, that the  linear system in~\eqref{nkrr} can be seen to obtained from the one in ~\eqref{krr} by uniform column subsampling of the empirical kernel matrix. The columns selected corresponds to the inputs $\{\widetilde{x}_1, \dots, \widetilde{x}_M\}$. FALKON proposes to compute a solution of the linear system~\ref{nkrr} via a preconditioned iterative solver. The preconditioner is the core  of the algorithm and is defined by a matrix $B$ such that 
\be\label{prec}
BB^\top  = \left(\frac{n}{M}K_{MM}^2 + \la K_{MM}\right)^{-1}.
\ee
The above choice provides a computationally efficient approximation  to the exact preconditioner of the linear system in \eqref{nkrr} corresponding to    $B$ such that 
$
BB^\top  = (K_{nM}^\top K_{nM} + \la K_{MM})^{-1}.
$
The preconditioner in~\eqref{prec} can then be combined with conjugate gradient to solve the linear system in~\eqref{nkrr}.  The overall algorithm has complexity  $\bigotime(n M t)$ in time and $\bigotime(M^2)$ in space, where $t$ is the number of conjugate gradient iterations performed.
\\
In this paper,  we analyze a variation of FALKON where the points $\{\widetilde{x}_1, \dots, \widetilde{x}_M \}$
are selected via leverage score sampling using BLESS, see \Cref{alg:est-ridge-lev-scores} or \Cref{alg:mahoneyplus-nystrom-fast}, so that  $M = M_h$ and $\widetilde{x}_k = x_{j_k}$, for $J_h = \{j_1,\dots, j_{M_h}\}$ and $k \in [M_h]$. Further, the  preconditioner in~\eqref{prec} is replaced by 
\eqal{\label{eq:FALKON-BLESS-prec}
B_h B_h^\top = \left(\frac{n}{M}K_{J_h, J_h} A_h^{-1}K_{J_h, J_h} + \la_h K_{J_h, J_h}\right)^{-1}.
}
This  solution can lead to  huge  computational improvements. Indeed, the total cost of FALKON-BLESS is the sum of computing BLESS and FALKON, corresponding to 
\be\label{FLKLSG-costs} 
O\left(n M t + (1/\la)M^2\log n + M^3\right)  \quad \quad \quad \bigotime(M^2), 
\ee 
in time and space respectively,  where $M$ is the size of the set $J_H$ returned by BLESS.

\begin{table}[t]
	\begin{tabular}{@{}cc@{}}
		\begin{minipage}[]{0.535\linewidth}
			\center
			\begin{tabular}{@{}lccccccc@{}}
				\toprule
				& Time  &  R-ACC & $5^{th} /~ 95^{th}$ quant \\
				\midrule
				BLESS & \textbf{17}      		& {1.06} 	    	&{0.57} / {2.03} \\
				BLESS-R & \textbf{17}      	& {1.06}  		&{0.73} / {1.50} \\
				SQUEAK & {52}      		& {1.06} 	       	&{0.70} / {1.48} \\
				Uniform & {-}  			& {1.09} 		&{0.22} / {3.75} \\
				RRLS & {235}    	 	& {1.59} 		&{1.00} / {2.70} \\
				\bottomrule
			\end{tabular}
		\end{minipage}
		&
		\hspace{-.5cm}
		\begin{minipage}[]{0.470\linewidth}
			\center
			\includegraphics[height=5cm]{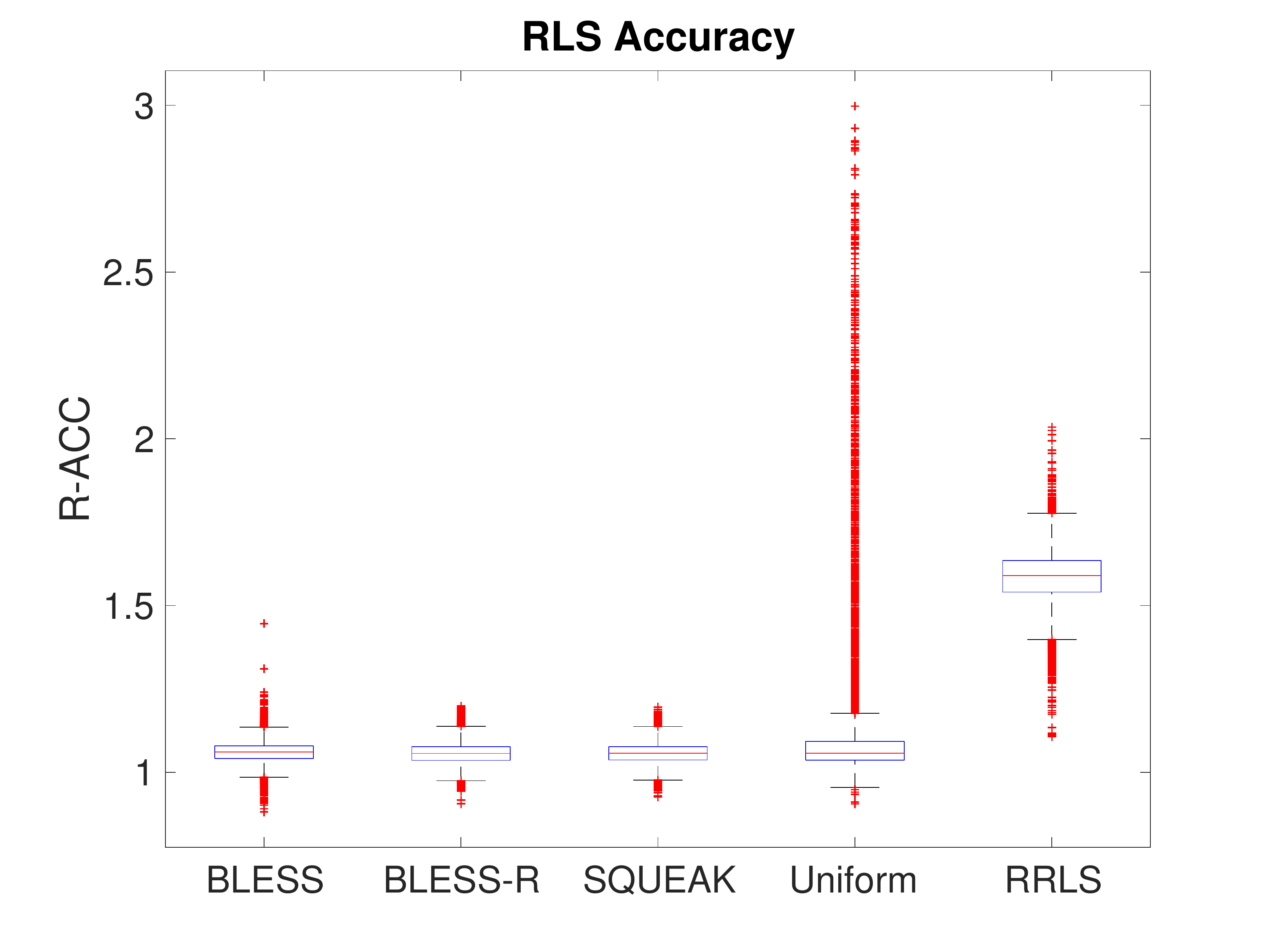}
		\end{minipage}
	\end{tabular}
	\captionof{figure}{\footnotesize Leverage scores relative accuracy for $\lambda = 10^{-5}, n = 70\,000, M=10\,000$, 10 repetitions.}\label{table1}
\end{table}

\subsection{Statistical properties  of FALKON-BLESS}
In this section,  we state and discuss our second main result, providing an excess risk bound for FALKON-BLESS.
Here a population version of the effective dimension plays a key role.  
Let $\rhox$ be the marginal measure of $\rho$ on $\X$, let $C:\hh \to \hh$ be the linear operator defined as follows and $\deff^*(\la)$ be the population version of $\deff(\la)$,
$$
\deff^*(\la)  =\tr(C(C+\la I)^{-1}), \quad  \textrm{with} \quad (C f)(x') = \int_\X K(x',x) f(x) d\rhox(x),
$$
for any $f \in \hh, x \in \X$.
It is possible to show that $\deff^*(\lambda)$ is the limit of $\deff(\lambda)$ as $n$ goes to infinity, see Lemma~\ref{lm:deff-to-deff*} below taken from  \cite{rudi2015less}. 
If we assume throughout that, 
\be\label{kerbound}
K(x,x')\le \kappa^2, \quad \forall x,x'\in \X,
\ee 
then the operator $C$ is symmetric,  positive definite and trace class, and the behavior of $\deff^*(\lambda)$ can be characterized in terms of the properties of the eigenvalues $(\sigma_j)_{j \in \N}$ of $C$. Indeed as for $\deff(\la)$, we have that $\deff^*(\la) \leq \kappa^2/\la$, moreover if $\sigma_j = \bigotime(j^{-\alpha})$, for
$\alpha \ge 1$, we have $\deff^*(\la) = \bigotime(\la^{-1/\alpha})$ . Then for larger $\alpha$, $\deff^*$ is smaller than $1/\la$ and faster learning rates are possible,  as shown below.
 \\
We next  discuss the properties of the  FALKON-BLESS solution denoted by  $\widehat{f}_{\la, n, t}$. 
\begin{theorem}\label{thm:FALKON-basic-rates}
	Let $n \in \N$, $\la > 0$ and $\delta \in (0,1]$. Assume that $y \in [-\frac{a}{2}, \frac{a}{2}]$, almost surely, $a > 0$, and denote by $f_\hh$ a minimizer of~\eqref{eq:learning-problem}. 
	 There exists $n_0 \in \N$,  such that for any $n \geq n_0$, if $t \geq \log n$, $\la \geq \frac{9\kappa^2}{n} \log \frac{n}{\delta}$, then the following holds with probability at least $1-\delta$:
	 $${\cal R}(\widehat{f}_{\la, n, t}) \leq  \frac{4a}{n} ~+~ 32\|f_\hh\|^2_\hh \left( \frac{a^2 \log^2 \frac{2}{\delta}}{n^2 \la}  + \frac{a ~\deff(\la)~\log \frac{2}{\delta}}{n} + \la\right).$$
	 In particular, when $\deff^*(\la)= \bigotime(\la^{-1/\alpha})$, for $\alpha \geq 1$,  by selecting $\la_* = n^{-\alpha/(\alpha + 1)}$, we have   
	 $${\cal R}(\widehat{f}_{\la_*, n, t}) \le c n^{-\frac{\alpha}{\alpha + 1}},$$
	 where $c$ is given explicitly in the proof.
\end{theorem}
\noindent We comment on the above result discussing the statistical and computational implications.
\\
\\
{\bf Statistics.} The above theorem  provides statistical guarantees in terms of finite sample  bounds on the excess risk  of FALKON-BLESS,  A first bound depends of the number of examples $n$, the regularization parameter $\la$ and the population effective dimension $\deff^*(\la)$. The second bound is derived optimizing $\la$,  and is the same as the one achieved by exact kernel ridge regression which  is known to be optimal \cite{caponnetto,steinwart2009optimal,lin2018optimal}.  Note that improvements under further assumptions are  possible and are derived in the supplementary materials,  see Thm.~\ref{thm:generalization-FALKON-LSG}. Here, we comment  on the computational properties of FALKON-BLESS and compare it to previous solutions.
\\
\\
{\bf Computations.} To discuss computational implications, we recall  a result from  \cite{rudi2015less} showing that 
the population version of the effective dimension $\deff^*(\la)$
and the effective dimension $\deff(\la)$ associated to the empirical kernel matrix converge up to constants. 
\begin{lemma}\label{lm:deff-to-deff*}
	Let $\la > 0$ and $\delta \in (0,1]$. When $\la \geq \frac{9\kappa^2}{n} \log\frac{n}{\delta}$, then  with probability at least $1-\delta$,
	$$(1/3) \deff^*(\la) \leq \deff(\la) \leq 3 \deff^*(\la).$$
\end{lemma}
Recalling the complexity of FALKON-BLESS~\eqref{FLKLSG-costs}, using Thm~\ref{thm:FALKON-basic-rates} and \Cref{lm:deff-to-deff*}, we derive a cost
$$\bigotime\left(n \deff^*(\la) \log n  + \frac{1}{\la}\deff^*(\la)^2\log n + \deff^*(\la)^3\right)$$ 
in time and   $\bigotime(\deff^*(\la)^2)$ in space,  for all $n,\la$ satisfying the assumptions in Theorem~\ref{thm:FALKON-basic-rates}. These expressions can be further simplified. Indeed, it is easy to see that for all $\la>0$, 
\be\label{deffup}
\deff^*(\la)\le\kappa^2 /\la,
\ee so that $\deff^*(\la)^3\le\frac{\kappa^2}{\la}\deff^*(\la)^2$. Moreover,   if we consider the optimal choice  $\la_* = \bigotime(n^{-\frac{\alpha}{\alpha+1}})$ given in Theorem~\ref{thm:FALKON-basic-rates}, and take $\deff^*(\la) = \bigotime(\la^{-1/\alpha})$, we have $\frac{1}{\la_*} \deff^*(\la_*) \leq \bigotime(n)$, and therefore  $\frac{1}{\la}\deff^*(\la)^2 \leq \bigotime(n \deff^*(\la))$. In summary, for the parameter choices leading to optimal learning rates, 
FALKON-BLESS has complexity $\wt{\bigotime}(n \deff^*(\la_*) )$, in time and $\wt{\bigotime}(\deff^*(\la_*)^2)$ in space, ignoring log terms.
We can compare this to previous results. In \cite{rudi2017falkon} uniform sampling is considered  
leading to $M \leq \bigotime(1 / \la)$ and achieving a complexity of $\wt{\bigotime}(n/\la)$ which  is always larger than the one achieved by FALKON in view of~\eqref{deffup}.
Approximate leverage scores sampling is also considered in  \cite{rudi2017falkon} requiring $\wt{\bigotime}(n \deff(\la)^2)$ time and reducing the time complexity of FALKON to $\wt{\bigotime}(n\deff(\la_*))$. Clearly in this case the complexity of leverage scores sampling dominates, and our results provide BLESS as a fix.

  \begin{figure*}[t]
	\begin{minipage}{0.49\textwidth}
		\hspace{-.3cm}
		\centering
		\includegraphics[height=5cm]{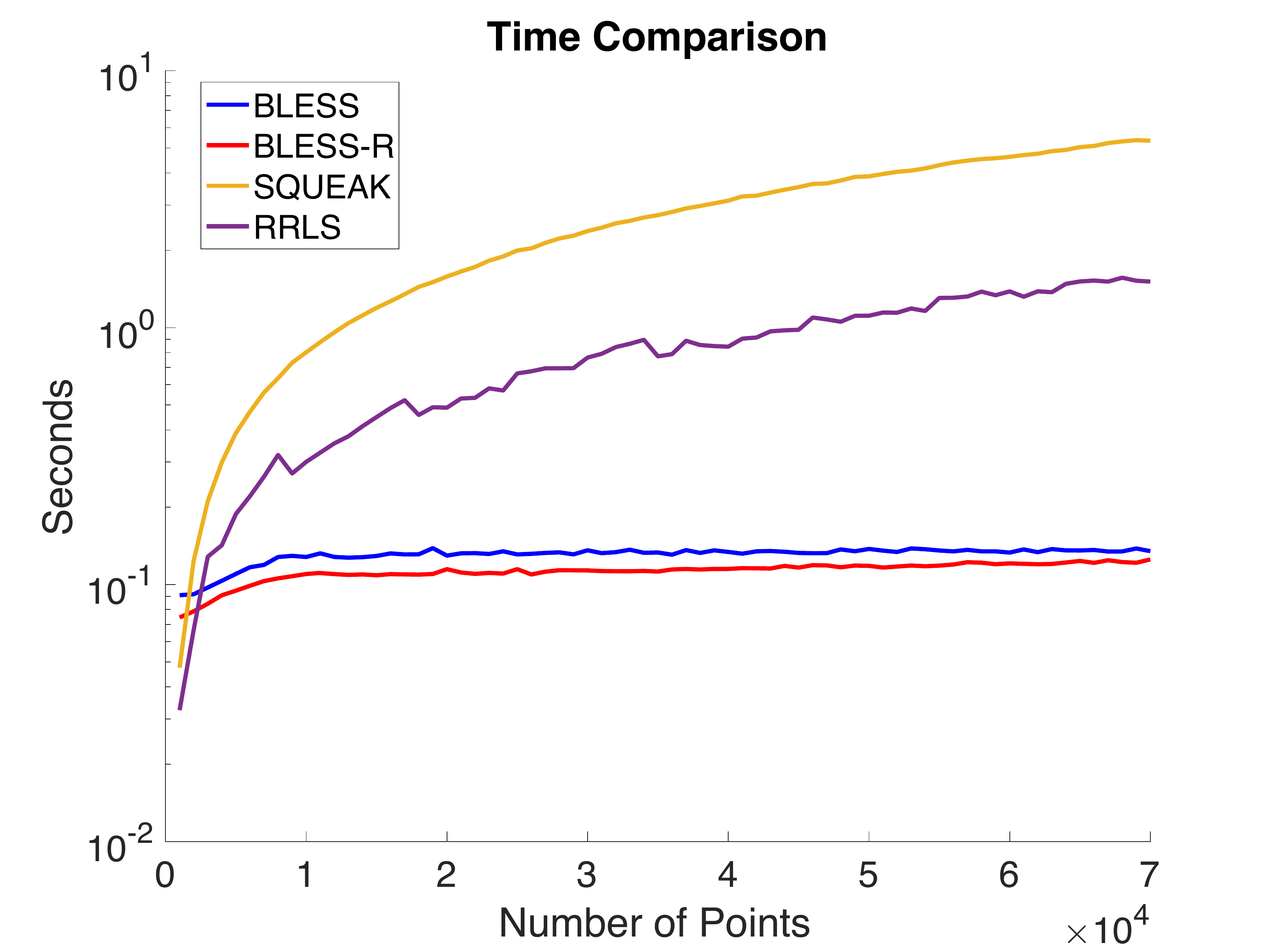}
		\captionof{figure}{\footnotesize Runtimes with $\la=10^{-3}$ and $n$ increasing}\label{fig:times}
	\end{minipage}
	\begin{minipage}{0.49\textwidth}
		\centering
		\hspace{-.3cm}
		\includegraphics[height=5cm]{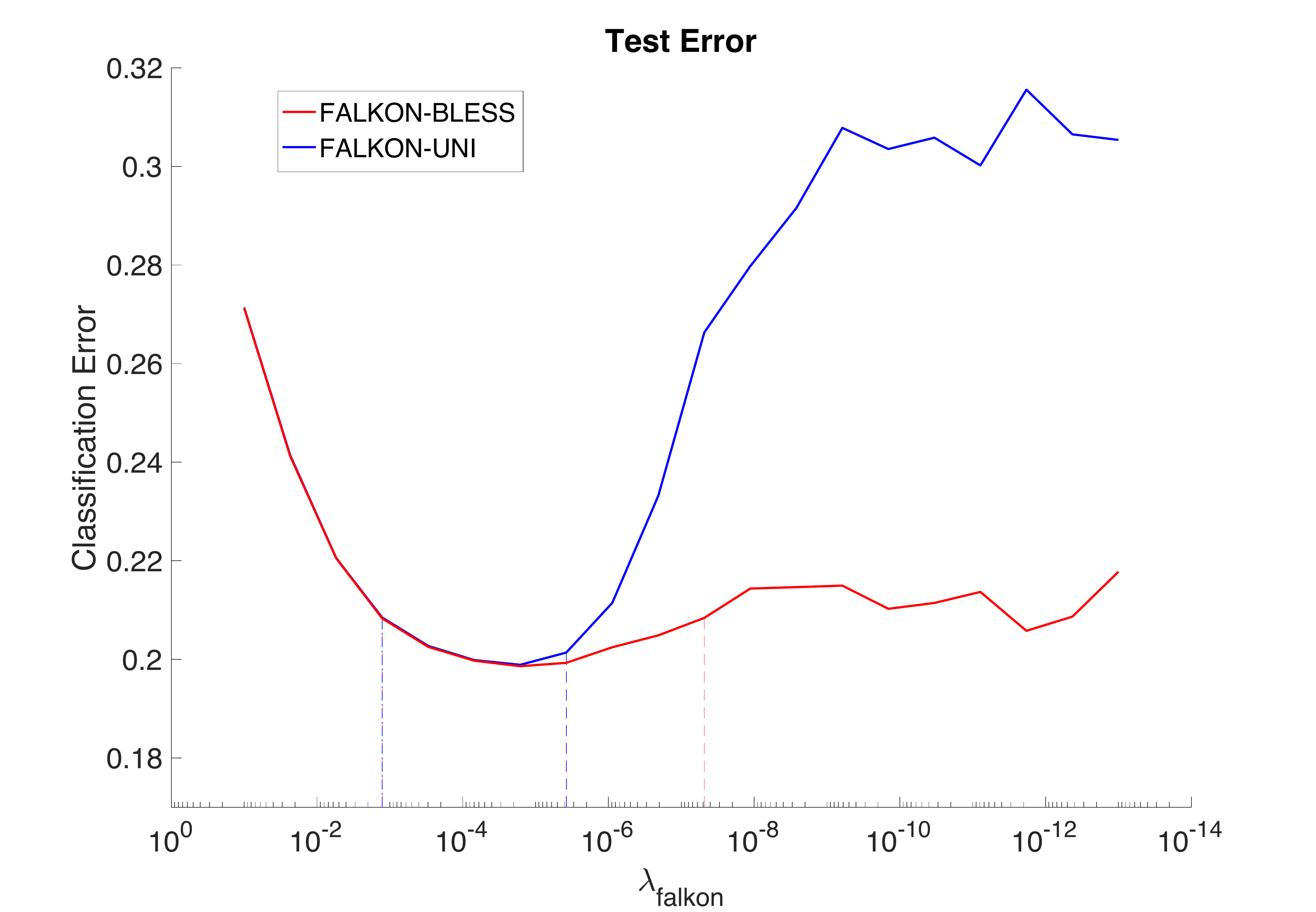}
		\captionof{figure}{\footnotesize C-err at 5 iterations for varying $\lambda_{falkon}$ }\label{fig:rob}
	\end{minipage}
\end{figure*}
\section{Experiments}

{\bf Leverage scores  accuracy.} 
We first study the accuracy of the leverage scores generated by BLESS and BLESS-R,
comparing SQUEAK \cite{calandriello_disqueak_2017} and Recursive-RLS (RRLS) \cite{musco2016provably}.
We begin by uniformly sampling a subsets of $n = 7 \times 10^4$ points from the SUSY dataset \cite{baldi2014searching},
and computing the exact leverage scores $\ell(i,\lambda)$ using a Gaussian Kernel with $\sigma = 4$ and $\la = 10^{-5}$,
which is at the limit of our computational
feasibility. We then run each algorithm to compute the  approximate
leverage scores $\alglev_{J_H}(i,\lambda)$, and we measure the accuracy of each method using
the ratio $\alglev_{J_H}(i,\lambda)/\ell(i,\lambda)$ (R-ACC).
The final results are presented in Figure~\ref{table1}.
On the left side for each algorithm we report runtime, mean R-ACC, and the $5^{th}$ and $95^{th}$ quantile, each averaged over the 10 repetitions.
On the right side a box-plot of the R-ACC.
As shown in Figure~\ref{table1} BLESS and BLESS-R achieve the same optimal accuracy of  SQUEAK 
with just a fraction of time.
Note that despite  our best efforts, we could not obtain high-accuracy results for RRLS (maybe a wrong constant in the original implementation). However note that RRLS is computationally demanding compared to BLESS, being orders of magnitude slower, as expected from the theory.
Finally, although uniform sampling is the fastest approach, it suffers from much larger variance and can over or under-estimate leverage scores by an order of magnitude more than the other methods, making it more fragile for downstream applications.\\
In Fig.~\ref{fig:times} we plot the runtime cost of the compared algorithms as the number of points grows from $n=1000$ to $70000$, this time for $\la = 10^{-3}$.
We see that while previous algorithms' runtime grows near-linearly with $n$, BLESS and BLESS-R run in a constant $1/\lambda$ runtime, as predicted by the theory.
\begin{figure*}[t]
\begin{minipage}{0.49\textwidth}
 \centering
 \includegraphics[height=5cm]{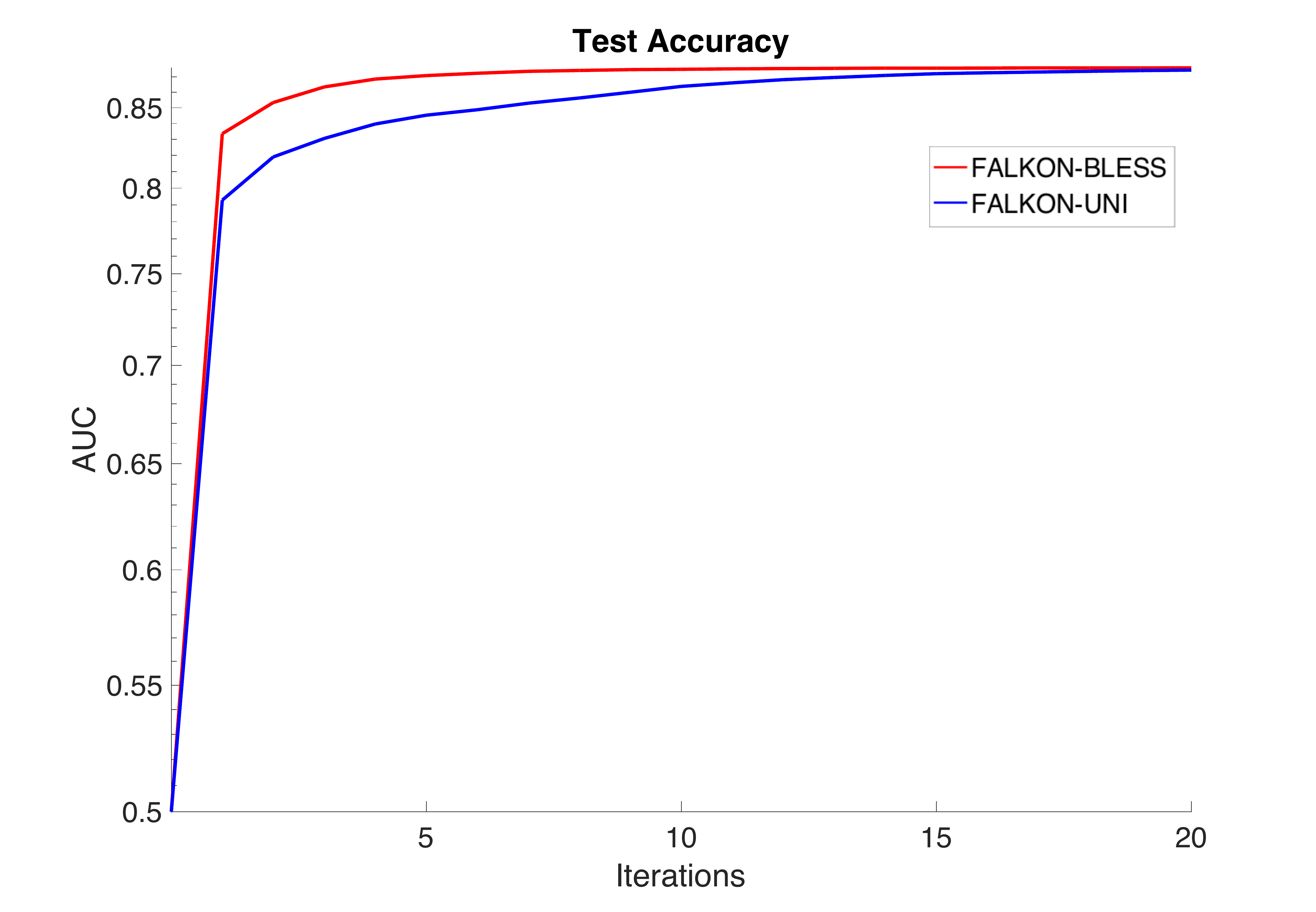}
 \captionof{figure}{\footnotesize AUC per iteration of the SUSY dataset}\label{fig1}
 \end{minipage}
 \begin{minipage}{0.49\textwidth}
 \centering
 \includegraphics[height=5cm]{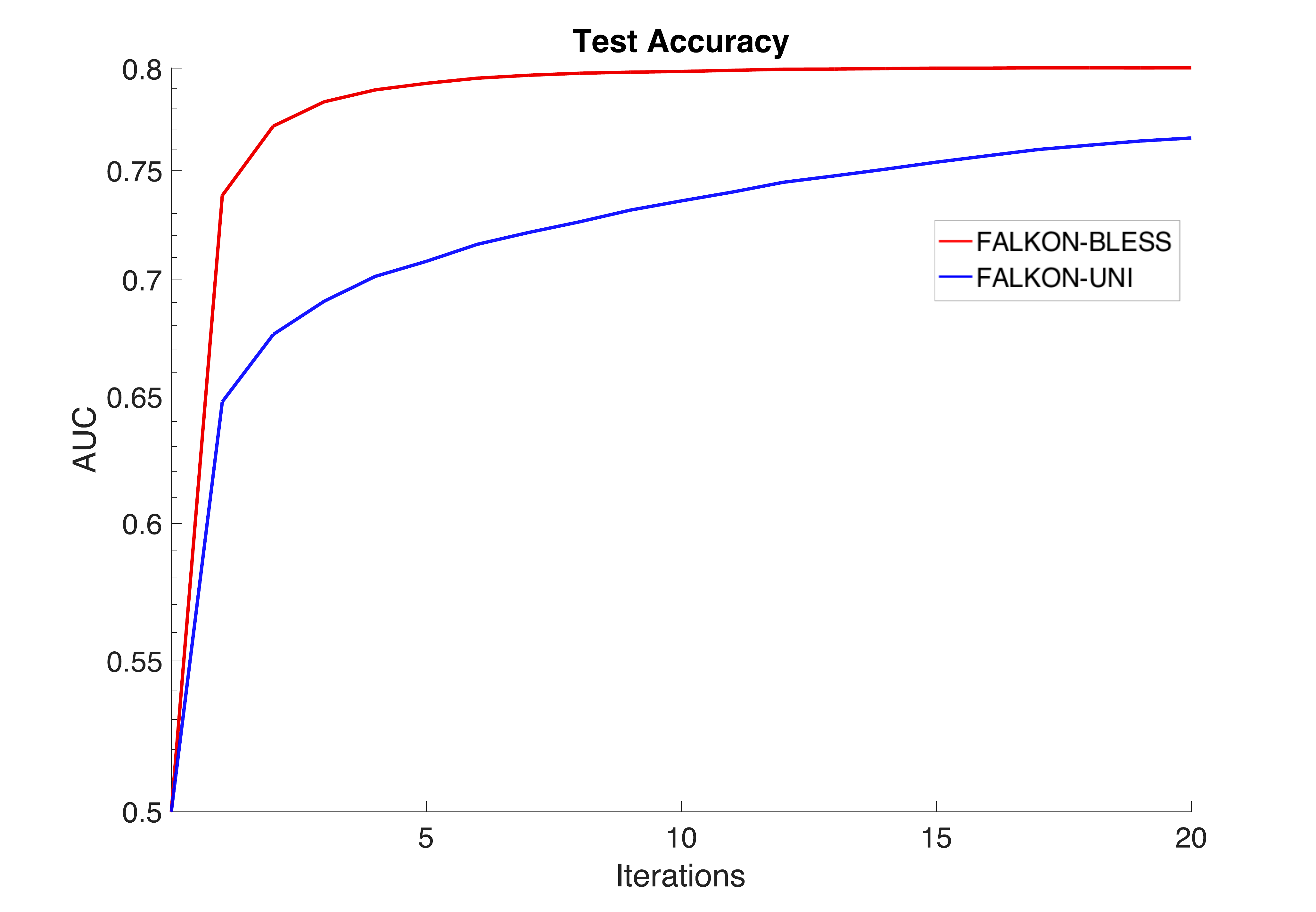}
 \captionof{figure}{\footnotesize AUC per iteration of the HIGGS dataset}\label{fig2}
\end{minipage}
\end{figure*}
\\
{\bf BLESS for supervised learning.}
We study the performance of FALKON-BLESS  and compare it with the original FALKON \cite{rudi2017falkon} where an equal number of \Nystrom~ centres are sampled uniformly at random (FALKON-UNI).
We take from \cite{rudi2017falkon} the two biggest datasets and their best hyper-parameters for the FALKON algorithm.
\\
We noticed that it is possible to achieve the same accuracy of FALKON-UNI, by using $\lambda_{bless}$ for BLESS and   $\lambda_{falkon}$ for FALKON with $\la_{bless} \gg \la_{falkon}$, in order to lower the $\deff$ and keep the number of \Nystrom~ centres low.
For the SUSY dataset we use a Gaussian Kernel with $\sigma = 4, \lambda_{falkon} = 10^{-6}, \lambda_{bless} = 10^{-4}$ obtaining $M_H \simeq 10^4$ \Nystrom~ centres.
For the HIGGS dataset we use a Gaussian Kernel with $\sigma = 22, \lambda_{falkon} = 10^{-8}, \lambda_{bless} = 10^{-6}$, obtaining $M_H \simeq 3\times10^4$ \Nystrom~ centres. We then sample a comparable number of centers uniformly for FALKON-UNI.
Looking at the plot of their AUC at each iteration (Fig.\ref{fig1},\ref{fig2}) we observe that  FALKON-BLESS converges much faster than FALKON-UNI.
For the SUSY dataset (Figure~\ref{fig1}) 5 iterations of FALKON-BLESS (160 seconds) achieve the same accuracy of 20 iterations of FALKON-UNI (610 seconds). Since running BLESS takes just $12$ secs. this corresponds to a $\sim4\times$ speedup.  For the HIGGS dataset 10 iter. of FALKON-BLESS (with BLESS requiring $1.5$ minutes, for a total of $1.4$ hours) achieve better accuracy of 20 iter. of FALKON-UNI ($2.7$ hours).
Additionally we observed that FALKON-BLESS is more stable than FALKON-UNI w.r.t. $\la_{falkon}, \sigma$.
In Figure~\ref{fig:rob} the classification error after 5 iterations of FALKON-BLESS and FALKON-UNI over the SUSY dataset ($\la_{bless}=10^{-4}$). We notice that FALKON-BLESS has a wider optimal region ($95\%$ of the best error) for the regulariazion parameter ($\small[1.3\times 10^{-3}, 4.8 \times 10^{-8}]$) w.r.t. FALKON-UNI ($\small[1.3 \times 10^{-3},3.8 \times 10^{-6}]$).

\section{Conclusions}
In this paper we presented two algorithms BLESS and BLESS-R to efficiently compute a small set of columns from a large symmetric positive semidefinite matrix $K$, useful for approximating the matrix or to compute leverage scores with a given precision. Moreover we applied the proposed algorithms in the context of statistical learning with least squares, combining BLESS with FALKON \cite{rudi2017falkon}. We analyzed the computational and statistical properties of the resulting algorithm, showing that it achieves optimal statistical guarantees with a cost that is $O(n \deff^*(\la))$ in time,  being currently the fastest. We can extend the proposed work in several ways: (a) combine BLESS with fast stochastic \cite{roux2012stochastic} or online \cite{NIPS2017_7194} gradient algorithms  and other approximation schemes (i.e. random features \cite{rahimi2008random, rudi2017generalization, NIPS2018_8222}), to further  reduce the computational complexity for optimal rates, (b) consider the impact of BLESS  in the context of  multi-tasking \cite{argyriou2008convex,ciliberto2017consistent} or structured prediction \cite{ciliberto2016,korba2018structured}.

\vspace{.6cm}
{\noindent\small	
{\bf Acknowledgments.}$ $\\
This material is based upon work supported by the Center for Brains, Minds and Machines (CBMM), funded by NSF STC award CCF-1231216, and the Italian Institute of Technology.  We gratefully acknowledge the support of NVIDIA Corporation for the donation of the Titan Xp GPUs and the Tesla k40 GPU used for this research.
L.~R. acknowledges the support of the AFOSR projects FA9550-17-1-0390  and BAA-AFRL-AFOSR-2016-0007 (European Office of Aerospace Research and Development), and the EU H2020-MSCA-RISE project NoMADS - DLV-777826. A. R. acknowledges the support of the European Research Council (grant SEQUOIA 724063).
}
\bibliographystyle{unsrt}
\bibliography{biblio}

\newpage
\appendix

 \section{Theoretical Analysis for \Cref{alg:est-ridge-lev-scores,alg:mahoneyplus-nystrom-fast} }
 
 In this section, Thm.~\ref{thm:alg-appr-lev-scores-extended-form} and Thm.~\ref{thm:alg-mplus-extended-form} provide guarantees for the two methods, from which Thm.~\ref{thm:main-appr-lev-scores} is derived.
 
 In particular in Section~\ref{sect:analytic-dec} some important properties about (out-of-sample-)leverage scores, that will be used in the proofs, are derived.

 \subsection{Notation}
 Let $\X$ be a Polish space and $K:\X\times\X\to\R$ a positive semidefinite function on $\X$, we denote $\hh$ the Hilbert space obtained by the completion of 
 \eqals{
 	\hh = \overline{\textrm{span} \{ K(x,\cdot) \ | \ x\in\X\} }
 }
 according to the norm induced by the inner product $\scal{K(x,\cdot)}{K(x',\cdot)}_\hh = K(x,x')$. Spaces $\hh$ constructed in this way are known as {\it reproducing kernel Hilbert spaces} and there is a one-to-one relation between a kernel $K$ and its associated RKHS. For more details on RKHS we refer the reader to \cite{aronszajn1950theory,steinwart2008support}. Given a kernel $K$, in the following we will denote with $K_x = K(x,\cdot) \in\hh$ for all $x\in\X$. We say that a kernel is bounded if $\|K_x\|_\hh \leq \kappa$ with $\kappa>0$.  In the following we will always assume $K$ to be continuous and bounded by $\kappa>0$. The continuity of $K$ with the fact that $\X$ is Polish implies $\hh$ to be separable \cite{steinwart2008support}.
 
 In the rest of the appdendizes we denote with $A_\la$, the operator $A + \la I$, for any symmetric linear operator $A$, $\la \in \R$ and $I$ the identity operator.

 \subsection{Definitions}

 For $n\in \N$, $(x_i)_{i=1}^n$, and $J \subseteq \{1,\dots, n\}$, $A \in \R^{|J|\times |J|}$ diagonal matrix with positive diagonal, denote $\alglev_J$ in \cref{def:out-of-sample-lev-scores} by showing the dependence from both $J$ and $A$ as
 \be\label{def:out-of-sample-lev-scores-general}
 \alglev_{J,A}(i,\lambda) =(\la n)^{-1} (\wh{K}_{ii} -  \wh{K}_{J,i}^\top (\wh{K}_{J,J} + \lambda n A)^{-1} \wh{K}_{J,i}).
 \ee
 Moreover define $\widehat{C}_{J,A}$ as follows
 $$\widehat{C}_{J, A} = \frac{1}{|J|} \sum_{i =1}^{|J|} A_{ii}^{-1} K_{x_{j_i}} \otimes K_{x_{j_i}}.$$
 We define the {\em out-of-sample leverage scores}, that are an extension of $\alglev_{J,A}$ to any point $x$ in the space $\X$.
 \bd[out-of-sample leverage scores]\label{def:out-of-sample-lev-scores-true}
 Let $J = \{j_1,\dots,j_M\} \subseteq \{1,\dots, n\}$, with $M \in \mathbb{N}$ and $A \in \mathbb{R}^{M \times M}$ be a positive diagonal matrix. Then for any $x \in X$ and $\lambda > 0$ we define
 $$\emplev_{J,A}(x, \la) =  \frac{1}{n}\|(\widehat{C}_{J,A} + \la I)^{-1/2}  K_x\|^2_\hh.$$
 Moreover define $\emplev_{\emptyset, []} (x, \la) = (\la n)^{-1} K(x,x)$.
 \ed
 In particular we denote by
 $$\emplev(x, \la) = \emplev_{[n], I}(x,\la),$$
 the out of sample version of the leverage scores $\ell(i, \la)$. Indeed note that $\emplev(x_i,\la) = \ell(i,\la)$ for $i \in [n]$ and $\la >0$ as proven by the next proposition that shows, more generally, the relation between $\emplev_{J,A}$ and $\alglev_{J,A}$.
 
 \bp\label{prop:char-emplev-op}
 Let $n \in \N$, $(x_i)_{i=1}^n \subseteq X$. For any $\la > 0, J \subseteq \{1,\dots, n\}, A \in \R^{|J| \times |J|}$ with $A$ positive diagonal, we that that for any $x \in X$, $\emplev_{J,A}(x, \la)$ in Def.~\ref{def:out-of-sample-lev-scores-true} and $\alglev_{J, A}(x,\la)$ in Def.~\ref{def:out-of-sample-lev-scores}, satisfy
 $$\emplev_{J,\frac{n}{|J|}A}(x_i, \la) =  \alglev_{J,A}(i,\la),$$
 when $|J| > 0$, and $\emplev_{\emptyset, []}(x_i, \la) =  \alglev_{\emptyset, []}(i,\la)$, when $|J| = 0$, for any $i \in [n], \la > 0$.
 \ep
 \bpr
 Let $J = \{j_1,\dots, j_{|J|}\}$.  We will first show that $\emplev_{J,A}(x,\lambda)$ is characterized by,
 $$\emplev_{J,A}(x,\lambda) = \frac{1}{\la n}K(x,x) - \frac{1}{\la n}v_J(x)^\top (K_J + \lambda |J| A)^{-1} v_J(x),$$
 with $K_J \in \mathbb{R}^{M \times M}$ with $(K_J)_{lm} = K(x_{j_l}, x_{j_m})$ and $v_J(x) = (K(x,x_{j_1}),\dots, K(x,x_{j_M}))$.
 Denote with $Z_{J}: \hh \to \R^{|J|}$, the linear operator defined by $Z_J = (K_{x_{j_1}}, \dots, K_{x_{j_{|J|}}})^\top$, that is $(Z_J f)_k = \big< K_{x_{j_k}}, f\big>_\hh$, for $f \in \hh$ and $k \in \{1,\dots |J|\}$. Then, by denoting with $B = |J|A$ we have
 $$Z_J^* B^{-1} Z_J = \frac{1}{|J|} \sum_{i=1}^{|J|} A_{ii}^{-1} K_{x_{j_i}} \otimes K_{x_{j_i}} = \widehat{C}_{J,A}.$$
 Now note that, since $(Q + \la I)^{-1} = \la^{-1}(I - Q(Q+\la I)^{-1})$ for any positive linear operator and $\la > 0$, we have 
 \eqals{
 	\emplev_{J,A}(x, \la) &= \frac{1}{n} \scal{K_x}{(\widehat{C}_{J,A} + \la I)^{-1} K_x}_\hh = \frac{1}{\la n} \scal{K_x}{(I - \widehat{C}_{J,A}(\widehat{C}_{J,A} + \la I)^{-1}) K_x}_\hh \\
 	& = \frac{K(x,x)}{\la n} - \frac{1}{\la n} \scal{K_x}{Z_J^* B^{-1/2}(B^{-1/2} Z_J Z_J^*B^{-1/2} + \la I)^{-1}B^{-1/2} Z_J K_x}_\hh,
 }
 where in the last step we use the fact that $R^*R(R^*R + \la I)^{-1} = R^*(R R^* + \la I)^{-1}R$, for any bounded linear operator $R$ and $\la > 0$. In particular we used it with $R = B^{-1/2} Z_J$. Now note that $Z_J Z_J^* \in \R^{|J| \times |J|}$ and in particular $Z_J Z_J^* = K_J$, moreover $Z_J K_x = v(x)$, so
 \eqals{
 	\emplev_{J,A}(x, \la) &= \frac{K(x,x)}{\la n} - \frac{1}{\la n} v(x)^\top B^{-1/2}(B^{-1/2} K_J B^{-1/2} + \la I)^{-1}B^{-1/2} v(x) \\
 	&= \frac{K(x,x)}{\la n} - \frac{1}{\la n} v(x)^\top ( K_J + \la B)^{-1} v(x) \\
 	&= \frac{K(x,x)}{\la n} - \frac{1}{\la n} v(x)^\top ( K_J + \la |J|A)^{-1} v(x),
 }
 where in the second step we used the fact that $B^{-1/2}(B^{-1/2} K B^{-1/2} + \la I)^{-1}B^{-1/2} = (K + \la B)^{-1}$, for any invertible $B$ any positive operator $K$ and $\la > 0$.
 
 Finally note that
 $$\emplev_{J,\frac{n}{|J|}A}(x_i, \la) = \frac{K(x,x)}{\la n} - \frac{1}{\la n} v(x)^\top ( K_J + \la n A)^{-1} v(x) = \alglev_{J,A}(i,\la).$$
 \epr

 \subsection{Preliminary results}
 
 Denote with $G_\la(A,B)$ the quantity
 $$G_\la(A,B)  = \|(A + \la I)^{-1/2}(A - B)(A + \la I)^{-1/2}\|,$$
 for $A, B$ positive bounded linear operators and for $\la > 0$. 
 
 \bp\label{prop:AlaBla}
 Let $A, B$ be positive bounded linear operators and $\la > 0$,  then
 $$\| I - (A + \la I)^{-1/2} (B + \la I) (A + \la I)^{-1/2}\|  = G_\la(A, B) \leq \frac{G_\la(B,A)}{1-G_\la(B,A)},$$
 where the last inequality holds if $G_\la(B,A) < 1$.
 \ep
 \bpr
 For the sake of compactness denote with $A_\la$ the operator $A + \la I$ and with $B_\la$ the operator $B + \la I$.
 First of all note that $I = A_\la^{-1/2}  A_\la A_\la^{-1/2}$, so
 \eqals{
 	I - A_\la^{-1/2} B_\la A_\la^{-1/2} &= A_\la^{-1/2}  A_\la A_\la^{-1/2} - A_\la^{-1/2} B_\la A_\la^{-1/2} \\
 	& = A_\la^{-1/2} (A_\la - B_\la) A_\la^{-1/2} = A_\la^{-1/2} (A - B) A_\la^{-1/2}\\
 	& =A_\la^{-1/2} B_\la^{1/2}~~ B_\la^{-1/2}(A - B) B_\la^{-1/2} ~~ B_\la^{1/2} A_\la^{-1/2},
 }
 where in the last step we multiplied and divided by $B_\la^{1/2}$. Then
 $$\nor{I - A_\la^{-1/2} B_\la A_\la^{-1/2}} \leq \|A_\la^{-1/2} B_\la^{1/2}\|^2 \|B_\la^{-1/2}(A - B) B_\la^{-1/2}\|,$$
 moreover, by Prop.~7 of \cite{rudi2015less} (see also Prop.~8 of \cite{rudi2017generalization}), if $G_\la(B,A) < 1$, we have
 $$\|A_\la^{-1/2} B_\la^{1/2}\|^2 \leq (1- G_\la(B,A))^{-1}.$$
 \epr
 
 \bp\label{prop:split-diff}
 Let $A, B, C$ be bounded positive linear operators on a Hilbert space. Let $\la > 0$. 
 Then, the following holds
 $$G_\la(A, C) \leq G_\la(A,B) + (1 + G_\la(A,B)) G_\la(B, C).$$
 \ep
 \bpr
 In the following we denote with $A_\la$ the operator $A + \la I$ and the same for $B, C$. Then
 $$\|A_\la^{-1/2} (A - C)A_\la^{-1/2}\| \leq \|A_\la^{-1/2} (A - B)A_\la^{-1/2}\| + \|A_\la^{-1/2} (B - C)A_\la^{-1/2}\|.$$
 Now note that, by dividing and multiplying for $B_\la^{1/2}$, we have
 \eqals{
 	\|A_\la^{-1/2} (B - C)A_\la^{-1/2}\| &= \|A_\la^{-1/2}B_\la^{1/2}~B_\la^{-1/2} (B - C)B_\la^{-1/2}B_\la^{1/2}A_\la^{-1/2}\| \\
 	& \leq \|A_\la^{-1/2}B_\la^{1/2}\|^2\|B_\la^{-1/2} (B - C)B_\la^{-1/2}\| = \|A_\la^{-1/2}B_\la^{1/2}\|^2 G_\la(B, C).
 }
 Finally note that, since $\nor{Z}^2 = \nor{Z^*Z}$ for any bounded linear operator $Z$, we have
 \eqals{
 	\|A_\la^{-1/2}B_\la^{1/2}\|^2 &= \|A_\la^{-1/2}B_\la A_\la^{-1/2}\| =  \|I + (I - A_\la^{-1/2}B_\la A_\la^{-1/2})\| \leq 1 + \| I - A_\la^{-1/2}B_\la A_\la^{-1/2}\|.
 }
 Moreover, by Prop.~\ref{prop:AlaBla}, we have that 
 \eqals{
 	\| I - A_\la^{-1/2}B_\la A_\la^{-1/2}\| = G_\la(A, B).
 }
 \epr
 
 \bp\label{prop:sigma-min-max}
 Let $B$ be a bounded linear operator, then
 $$1 - \| I -  BB^*\| \leq \sigma_{\min}(B)^2   \leq \sigma_{\max}(B)^2 \leq 1 + \| I -  BB^*\|.$$
 \ep
 \bpr
 Now we recall that, denoting by $\preceq$ the Lowner partial order, for a positive bounded operator $A$ such that $a I \preceq A \preceq b I$ for $0 \leq a  \leq b$, we have $(1 - b) I \preceq I - A \preceq (1 - a)I \preceq (1 + b) I $ and so, since $ B B^* = I - (I -  BB^*)$, we have
 $$ (1 - \| I -  BB^*\|) I \preceq \sigma_{\min}(B)^2 I \preceq B B^* \preceq \sigma_{\max}(B)^2 I \preceq 1 + (1 + \| I -  BB^*\|) I,$$
 from we have the desired result.
 \epr

  Let $\nor{\cdot}_{HS}$ denote the Hilbert-Schmidt norm.
 
 We recall and adapt to our needs a result from Prop.~8 of \cite{rudi2015less}.
 \bp\label{prop:concentration-ClaC}
 Let $\la > 0$ and $v_1,\dots,v_n$ with $n \geq 1$, be identically distributed random vectors on separable Hilbert space $\hh$, such that there exists $\kappa^2 > 0$ for which $\|v\|_\hh \leq \kappa^2$ almost surely. Denote by $Q$ the Hermitian operator $Q =  \frac{1}{n}\sum_{i=1}^{n} \expect{v_i \otimes v_i}$.
 Let $Q_n = \frac{1}{n}\sum_{i=1}^{n} v_i \otimes v_i$.  Then for any $\delta \in (0,1]$, the following holds
 $$
 \nor{(Q+\lambda I)^{-1/2}(Q-Q_n)(Q+\lambda I)^{-1/2}}{} \leq \frac{4\kappa^2 \beta}{3 \la n} + \sqrt{\frac{2\kappa^2\beta}{\la n}}$$
 with probability $1-\delta$ and $\beta = \log \frac{4\tr(Q(Q+\la I)^{-1})}{\nor{Q(Q+\la I)^{-1}}{}\delta} \leq  \frac{8\kappa^2(1+ \tr (Q_\la^{-1}Q))}{\|Q\|\delta}$.
 \ep
 \bpr
 Let $Q_\la = Q + \la I$.
 Here we apply non-commutative Bernstein inequality like \cite{tropp2012user} (with the extension to separable Hilbert spaces as in\cite{rudi2015less}, Prop.~12) on the random variables $Z_i = M - Q_\la^{-1/2}v_i \otimes Q_\la^{-1/2}v_i$ with $M_i = Q_\la^{-1/2}(\expect{v_i \otimes v_i})Q_\la^{-1/2}$ for $1 \leq i \leq n$. Note that the expectation of $Z_i$ is $0$.
 The random vectors are bounded by
 \eqals{
 	\|Q_\la^{-1/2}v_i \otimes Q_\la^{-1/2}v_i - M_i\| &= \|\mathbb{E}_{v'_i}[Q_\la^{-1/2}v'_i \otimes Q_\la^{-1/2}v'_i - Q_\la^{-1/2}v_i \otimes Q_\la^{-1/2}v_i]\|_\hh \\
 	&\leq 2\|\kappa^2\|\|(Q+\la)^{-1/2}\|^2 \leq \frac{2\kappa^2}{\la},
 }
 and the second orded moment is
 \begin{align*}
 \mathbb{E} (Z_i)^2 &= \mathbb{E} \;\;\scal{v_i}{Q_\la^{-1} v_i} \;Q_\la^{-1/2}v_i \otimes Q_\la^{-1/2}v_i \;\;\;-\;\;\; Q_\la^{-2}Q^2 \\
 & \leq \frac{\kappa^2}{\la} \expect{Q_\la^{-1/2}v_1 \otimes Q_\la^{-1/2}v_1} = \frac{\kappa^2}{\la} Q(Q+\la I)^{-1} =: S.
 \end{align*}
 Now we can apply the Bernstein inequality with {\em intrinsic dimension} in \cite{tropp2012user} (or  Prop.~12 in \cite{rudi2015less}). Now some considerations on $\beta$.
 It is $\beta = \log \frac{4\tr S}{\nor{S}{}\delta} = \frac{4\tr Q_\la^{-1}Q}{\nor{Q_\la^{-1}Q}{}\delta}$, now we need a lower bound for $\nor{Q_\la^{-1}Q}{} = \frac{\sigma_1}{\sigma_1 + \la}$ where $\sigma_1 = \nor{Q}$ is the biggest eigenvalue of $Q$, now, when $0 <\la \leq  \sigma_1$ we have $\beta \leq \frac{8\tr Q}{\la\delta}$.
 
 When $\la \geq \sigma_1$, note that $\tr (Q(Q+\la I)^{-1}) \leq \la^{-1} \tr (Q) \leq \kappa^2/\la$, then
 $$\frac{\tr (Q(Q+\la I)^{-1}) }{\nor{Q_\la^{-1}Q}} \leq \frac{\kappa^2}{\la \frac{\sigma_1}{\sigma_1 + \la}} = \frac{\kappa^2}{\la} + \frac{\kappa^2}{\sigma_1} \leq \frac{2\kappa^2}{\sigma_1}.$$
 
 So finally $\beta \leq \frac{8(\kappa^2/\|Q\| + \tr (Q_\la^{-1}Q))}{\delta}$
 \epr

 \subsection{Analytic decomposition}\label{sect:analytic-dec}
 
 \blm\label{lm:J1A1toJA}
 Let $\la > 0$, $J, J' \subseteq \{1,\dots, n\}$, with $|J|, |J'| \geq 1$ and $A \in \R^{|J| \times |J|}$, $A' \in \R^{|J'| \times |J'|}$ positive diagonal matrices, then
 $$ \frac{1 - 2\nu}{1-\nu} \emplev_{J',A'}(x,\la) \leq \emplev_{J,A}(x,\la) \leq \frac{1}{1-\nu} \emplev_{J',A'}(x,\la), \quad \forall x \in \X,$$
 with $\nu = G_\la(\widehat{C}_{J',A'}, \widehat{C}_{J,A})$.
 \elm
 \bpr
 By denoting with $B$ the operator
 \eqals{
 	B = (\widehat{C}_{J,A} + \la I)^{-1/2} (\widehat{C}_{J',A'} + \la I)^{1/2},
 }
 and according to the characterization of $\emplev_{J,A}(x, \la)$ via Prop.~\ref{prop:char-emplev-op}, we have
 \eqals{
 	\emplev_{J,A}(x, \la) &= n^{-1}\nor{(\widehat{C}_{J,A} + \la I)^{-1/2}  K_x}^2_\hh = n^{-1}\nor{B ~ (\widehat{C}_{J',A'} + \la I)^{-1/2} K_x}^2_\hh.
 }
 So, by recalling the fact that, by definition of Lowner partial order $\preceq$, we have $a \|v\|^2 \leq \|Av\|^2 \leq b \|v\|^2$, for any vector $v$ and bounded linear operator such that $ a I \preceq A^*A \preceq b I$ with $0 \leq a \leq b$, and the fact that $\sigma(A^*A) = \sigma(AA^*) = \sigma(A)^2$, we have
 $$  \sigma_{\min}(B)^2 \nor{(\widehat{C}_{J',A'}+ \la I)^{-1/2} K_x}^2_\hh \leq \nor{B(\widehat{C}_{J',A'}+ \la I)^{-1/2} K_x}^2_\hh \leq \sigma_{\max}(B)^2 \nor{(\widehat{C}_{J',A'}+ \la I)^{-1/2} K_x}^2_\hh.$$
 That, by Prop.~\ref{prop:char-emplev-op}, is equivalent to
 $$  \sigma_{\min}(B)^2 \emplev_{J',A'}(x, \la) \leq \emplev_{J,A}(x, \la) \leq \sigma_{\max}(B)^2 \emplev_{J',A'}(x, \la).$$
 By Prop.~\ref{prop:sigma-min-max} we have $1 - \| I -  BB^*\| \leq \sigma_{\min}(B)^2   \leq \sigma_{\max}(B)^2  \leq 1 + \| I -  BB^*\|$.
 Finally, by Prop.~\ref{prop:AlaBla}, we have
 $$\|I -  BB^*\| \leq \frac{\nu}{1 - \nu}.$$
 \epr
 
 \blm\label{lm:emplev-lala1}
 Let $0 < \la \leq \la'$, and $J \subseteq \{1,\dots, n\}$ and $A \in \R^{|J| \times |J|}$, then
 $$ \emplev_{J,A}(x, \la') \leq \emplev_{J,A}(x, \la) \leq \frac{\la'}{\la} \emplev_{J,A}(x, \la'), \quad \forall x \in \X.$$
 \elm
 \bpr
 If $|J| = 0$ we have that $\emplev_{\emptyset,[]}(x,\la) = \frac{K(x,x)}{\la n}$ and the desired result is easily verified. 
 If $|J| \geq 1$, let $B = (C_{J,A} + \la I)^{-1/2} (C_{J,A} + \la' I)^{1/2}$.
 By recalling the fact that, by definition of Lowner partial order $\preceq$, we have $a \|v\|^2 \leq \|Av\|^2 \leq b \|v\|^2$, for any vector $v$ and bounded linear operator such that $ a I \preceq A^*A \preceq b I$ with $0 \leq a \leq b$, and the fact that $\sigma(A^*A) = \sigma(AA^*) = \sigma(A)^2$, we have
 $$  \sigma_{\min}(B)^2 \nor{(\widehat{C}_{J,A}+ \la' I)^{-1/2} K_x}^2_\hh \leq \nor{B(\widehat{C}_{J,A}+ \la' I)^{-1/2} K_x}^2_\hh \leq \sigma_{\max}(B)^2 \nor{(\widehat{C}_{J,A}+ \la' I)^{-1/2} K_x}^2_\hh.$$
 That, by Prop.~\ref{prop:char-emplev-op}, is equivalent to
 $$  \sigma_{\min}(B)^2 \emplev_{J,A}(x, \la') \leq \emplev_{J,A}(x, \la) \leq \sigma_{\max}(B)^2 \emplev_{J,A}(x, \la').$$
 Now note that
 $$\sigma_{\min}(B)^2 \geq \inf_{\sigma \geq 0} \frac{\sigma + \la'}{\sigma + \la} =1, \quad \sigma_{\max}(B)^2 \geq \sup_{\sigma \geq 0} \frac{\sigma + \la'}{\sigma + \la} = \frac{\la'}{\la}.$$
 \epr

 \bt\label{thm:analytical-dec}
 Let $\la > 0$, $J \subseteq \{1,\dots, n\}$, with $|J| \geq 1$ and $A \in \R^{|J| \times |J|}$ positive diagonal. Then the following hold for any $x \in X$,
 $$ \frac{1 - 2\nu_{J,A}}{1 - \nu_{J,A}} \emplev(x, \la) \leq \emplev_{J,A}(x, \la) \leq \frac{1}{1-\nu_{J,A}} \emplev(x, \la),$$
 where $\nu_{J,A} = G_\la(\widehat{C}, \widehat{C}_{J,A})$. Morever note that for any $|U| \subseteq \{1,\dots, n\}$, we have
 \eqals{
 	\nu_{J,A} \leq \eta_U + (1+ \eta_U)\beta_{J,A,U},
 }
 with $\beta_{J,A, U} = G_\la(\widehat{C}_{U,I}, \widehat{C}_{J,A})$ and $\eta_{U} = G_\la(\widehat{C}, \widehat{C}_{U,I})$.
 \et
 \bpr
 By applying Lemma~\ref{lm:J1A1toJA}, with their $J' = \{1,\dots, n\}, A' = I$, and recalling that $\emplev(x,\la) = \emplev_{\{1,\dots, n\}, I}$, we have for all $x \in \X$
 $$ \frac{1 - 2\nu_{J,A}}{1-\nu_{J,A}} \emplev(x,\la) \leq \emplev_{J,A}(x,\la) \leq \frac{1}{1-\nu_{J,A}} \emplev(x,\la).$$
 To conclude the proof we bound $\nu_{J,A}$ in terms of $\beta_{J,A,U}$ and $\eta_{U}$, via Prop.~\ref{prop:split-diff}.
 \epr
 
 \subsection{Proof for \Cref{alg:est-ridge-lev-scores}}

 \blm\label{lm:concentrate-beta1}
 Let $n \in \N$, $(x_i)_{i=1}^n \subseteq \X$.
 Let $U \subseteq \{1,\dots n\}$, with $|U| \geq 1$. Let $(p_k)_{k=1}^{|U|} \subset \R$ be a non-negative sequence summing to $1$. Let $M \in \N$ 
 and $J = \{j_1,\dots, j_M\}$ with $j_i$ sampled i.i.d. from $\{1,\dots, |U|\}$ with probability $(p_k)_{k=1}^{|U|}$ and $A = |U|\textrm{diag}(p_{j_1}, \dots, p_{j_M})$. Let $\tau \in (0,1]$, and $s : = \sup_{k \in \{1,\dots,|U|\}} \frac{1}{|U|p_k}\|(\widehat{C}_{U,I} + \la I)^{-1/2} K_{x_{u_k}}\|_\hh^2$. When 
 $$M \geq 2 s \log\frac{4n}{\tau},$$
 then the following holds with probability at least $1-\tau$
 $$\|(\widehat{C}_{U,I} + \la I)^{-1/2}(\widehat{C}_{J, A} - \widehat{C}_{U,I})(\widehat{C}_{U,I} + \la I)^{-1/2}\| \leq \sqrt{\frac{4s\log\frac{4n}{\tau}}{M}}.$$
 \elm
 \bpr
 Denote with $\zeta_i$ the random variable 
 $$\zeta_i = \frac{1}{|U| p_k} (\widehat{C}_{U,I} + \la I)^{-1/2}(K_{x_{j_i}}\otimes K_{x_{j_i}})(\widehat{C}_{U,I} + \la I)^{-1/2},$$
 for $i \in \{1,\dots, M\}$. In particular note that $\zeta_1,\dots, \zeta_M$ are i.i.d. since $j_1,\dots,j_M$ are. Moreover note the following two facts 
 \eqals{
 	\|\zeta_i\| & = \sup_{k \in \{1,\dots,|U|\}} \frac{1}{|U|p_k}\|(\widehat{C}_{U,I} + \la I)^{-1/2} K_{x_{u_k}}\|_\hh^2 = s,\\
 	\expect{\zeta_i} &= \sum_{k=1}^{|U|} p_k~ \frac{1}{|U| p_k}(\widehat{C}_{U,I} + \la I)^{-1/2}(K_{x_k}\otimes K_{x_k})(\widehat{C}_{U,I} + \la I)^{-1/2} \\
 	& =  (\widehat{C}_{U,I} + \la I)^{-1/2}\widehat{C}_{U,I}(\widehat{C}_{U,I} + \la I)^{-1/2} =: W,
 }
 where for the second identity we used the fact that $d/l_k = 1/(p_k |U|)$.
 Since by definition of $\widehat{C}_{J,A}$ we have 
 \eqals{
 	\frac{1}{M} \sum_{i=1}^M \zeta_i & = (\widehat{C}_{U,I} + \la I)^{-1/2} \left(\frac{1}{|J|} \sum_{i=1}^M \frac{1}{A_{ii}} K_{x_{j_i}} \otimes K_{x_{j_i}}\right)(\widehat{C}_{U,I} + \la I)^{-1/2} \\
 	& =  (\widehat{C}_{U,I} + \la I)^{-1/2}\widehat{C}_{J,A}(\widehat{C}_{U,I} + \la I)^{-1/2},
 }
 then, by applying non-commutative Bernstein inequality (Prop.~\ref{prop:concentration-ClaC} is a version specific for our problem), we have
 \eqals{
 	\|(\widehat{C}_{U,I} + \la I)^{-1/2}(\widehat{C}_{J, A} - \widehat{C}_{U,I})(\widehat{C}_{U,I} + \la I)^{-1/2}\| & = \big\|\frac{1}{M}\sum_{i=1}^M(\zeta_i - \expect{\zeta_i})\big\| \leq \frac{2s\eta}{3 M} + \sqrt{\frac{2 s\|W\|\eta}{M}},
 }
 with probability at least $1-\tau$, and $\eta := \log \frac{4\tr(W)}{\tau\|W\|}$. In particular, by noting that $\|W\| \leq 1$ by definition, when $M \geq 2s\eta$, then
 $$\frac{2s\eta}{3 M} + \sqrt{\frac{2 s\|W\|\eta}{M}} \leq \frac{2s \eta}{3 M} + \sqrt{\frac{2 s\eta}{M}} \leq \frac{1}{3}\sqrt{\frac{2 s\eta}{M}} + \sqrt{\frac{2 s\eta}{M}}  \leq \sqrt{\frac{4s\eta}{M}}.$$
 To conclude note that $\frac{\tr(W)}{\|W\|} \leq \textrm{rank}(W) \leq |U| \leq n$, so $\eta \leq \log\frac{4n}{\tau}$.
 \epr
 
 \blm\label{lm:concentrate-beta2}
 Let $n, R \in \N$, $(x_i)_{i=1}^n \subseteq \X$.
 Let $U = \{u_1, \dots, u_R\}$ with $u_i$ i.i.d. with uniform probability on $\{1,\dots, n\}$. Let $\tau \in (0,1]$ and let $\la > 0$. When
 $$R \geq \frac{2n\kappa^2}{\la n+\kappa^2} \log \frac{4n}{\tau},$$
 then the following holds with probability $1-\tau$
 $$\|(\widehat{C}+\la I)^{-1/2}(\widehat{C}_{U,I} - \widehat{C})(\widehat{C}+\la I)^{-1/2}\| \leq \sqrt{\frac{4n\kappa^2\log \frac{4n}{\tau}}{(\la n +\kappa^2) R} }.$$
 \elm
 \bpr
 Denote by $\zeta_i$ the random variable  $\zeta_i = (\widehat{C}+\la I)^{-1/2}(K_{x_{u_i}} \otimes K_{x_{u_i}})(\widehat{C}+\la I)^{-1/2}$, for $i \in \{1,\dots, R\}$. Note that $\zeta_i$ are i.i.d. since $u_i$ are. Moreover note that 
 \eqals{
 	\|\zeta_i\| &= \sup_{i \in \{1,\dots, n\}} \|(\widehat{C}+\la I)^{-1/2} K_{x_i}\|^2 \leq \sup_{i \in \{1,\dots, n\}} \|(\frac{1}{n} K_{x_i} \otimes K_{x_i}+\la I)^{-1/2} K_{x_i}\|^2 \\
 	&\leq \frac{n\kappa^2}{\la n + \kappa^2} =: v.
 }
 Moreover note that
 $$\expect{\zeta_i} = \frac{1}{n} \sum_{i=1}^n (\widehat{C}+\la I)^{-1/2}(K_{x_i} \otimes K_{x_i})(\widehat{C}+\la I)^{-1/2} = (\widehat{C}+\la I)^{-1/2}\widehat{C}(\widehat{C}+\la I)^{-1/2} =: W.$$
 So we have, by non-commutative Bernstein inequality (Prop.~\ref{prop:concentration-ClaC} is a version specific for our problem),
 $$\|(\widehat{C}+\la I)^{-1/2}(\widehat{C}_{U,I} - \widehat{C})(\widehat{C}+\la I)^{-1/2}\| =\big\|\frac{1}{M}\sum_{i=1}^M(\zeta_i - \expect{\zeta_i})\big\| \leq
 \frac{2v\eta}{3 R} + \sqrt{\frac{2v\|W\|\eta}{ R}},
 $$
 with probability at least $1-\tau$, and $\eta := \log \frac{4\tr(W)}{\tau\|W\|}$. In particular, by noting that $\|W\| \leq 1$ by definition, when $R \geq \frac{2n\kappa^2\eta}{(\la n + \kappa^2) R}$, analogously to the end of the proof of Lemma~\ref{lm:concentrate-beta1}, we have
 $\frac{2v\eta}{3 R} + \sqrt{\frac{2v\|W\|\eta}{R}}  \leq \sqrt{\frac{4n\kappa^2\eta}{(\la n + \kappa^2)R}}.$
 To conclude note that $\frac{\tr(W)}{\|W\|} \leq \textrm{rank}(W) \leq n$, so $\eta \leq \log\frac{4n}{\tau}$.
 \epr
 
 \blm\label{lm:concentrate-Fh}
 Let $n, R \in \N$, $(x_i)_{i=1}^n \subseteq \X$.
 Let $U = \{u_1, \dots, u_R\}$ with $u_i$ i.i.d. with uniform probability on $\{1,\dots, n\}$. Let $\tau \in (0,1]$ and let $\la > 0$. When
 $$R \geq \frac{16n\kappa^2}{\la n+\kappa^2} \log \frac{4n}{\tau},$$
 then the following holds with probability $1-\tau$
 $$\frac{n}{R} \sum_{i =1}^R \emplev(x_{u_i}, \la) < \max\left(5, \frac{6}{5} \deff(\la)\right).$$
 \elm
 \bpr
 First of all denote with $z_i$ the random variable $z_i = \frac{n}{R}\emplev(x_{u_i}, \la)$ and note that $(z_i)_{i=1}^R$ are $i.i.d.$ since $(u_i)_{i=1}^R$ are. Moreover, by the characterization of $\emplev(x,\la)$ via Prop.~\ref{prop:char-emplev-op}, we have 
 $$|z_i| \leq \sup_{k \in \{1,\dots, n\}} \|(\widehat{C} + \la I)^{-1/2} K_{x_k}\|^2 \leq \|(K_{x_k} \otimes K_{x_k}/n + \la I)^{1/2} K_{x_k}\|^2 \leq \frac{\kappa^2}{R(\kappa^2/n + \la)} =: v,$$
 moreover we have
 \eqals{
 	\expect{z_i} & = \expect{\tr((\widehat{C} + \la I)^{-1} (K_{x_{u_i}} \otimes K_{x_{u_i}}))} = \tr((\widehat{C} + \la I)^{-1} \expect{K_{x_{u_i}} \otimes K_{x_{u_i}}}) \\
 	& = \tr\left((\widehat{C} + \la I)^{-1} \sum_{k=1}^n \frac{1}{n} K_{x_k} \otimes K_{x_k}\right) =  \tr\left((\widehat{C} + \la I)^{-1} \widehat{C}\right) = \deff(\la).
 }
 So by applying Bernstein inequality, the following holds with probability at least $1-\tau$
 $$\left|\frac{n}{R} \sum_{i =1}^R \emplev(x_{u_i}, \la) - \deff(\la)\right| = \left| \frac{1}{R} \sum_{i=1}^R(z_i - \expect{z_i})\right| \leq \frac{2 v \log\frac{2}{\tau}}{3R} + \sqrt{\frac{2 v \deff(\la)\log\frac{2}{\tau}}{3R}}.$$
 So we have 
 $$\frac{n}{R} \sum_{i =1}^R \emplev(x_{u_i}, \la) \leq \deff(\la) + \left|\frac{n}{R} \sum_{i =1}^R \emplev(x_{u_i}, \la) - \deff(\la)\right| \leq \deff(\la) + \frac{2 v \log\frac{2}{\tau}}{3R} + \sqrt{\frac{2 v \deff(\la)\log\frac{2}{\tau}}{R}}.$$
 Now, if $\deff(\la) \leq 4$, since $R \geq 16v \log\frac{2}{\tau}$, we have that
 $$\deff(\la)  + \frac{2 v \log\frac{2}{\tau}}{3R} + \sqrt{\frac{2 v \deff(\la)\log\frac{2}{\tau}}{R}} \leq 4 + \frac{1}{24} + \sqrt{\frac{1}{2}} < 5.$$
 If $\deff(\la) > 4$, since $R \geq 16v \log\frac{2}{\tau}$, we have
 \eqals{
 	\deff(\la)  + \frac{2 v \log\frac{2}{\tau}}{3R} + \sqrt{\frac{2 v \deff(\la)\log\frac{2}{\tau}}{3R}} &\leq \left(1 + \frac{1}{24 \deff(\la)} + \sqrt{\frac{1}{8 \deff(\la)}}\right) \deff(\la) < \frac{6}{5} \deff(\la).
 }
 \epr
 
 \bt\label{thm:alg-appr-lev-scores-extended-form}
 Let $n \in \N$, $(x_i)_{i=1}^n \subseteq X$. Let $\delta \in (0,1]$, ~$t,q > 1$, $\la > 0$ and $H, d_h, \la_h, J_h, A_h, U_h$ as in Alg.~\ref{alg:est-ridge-lev-scores}. Let $\bar{A}_h = \frac{n}{|J|} A_h$ and $\nu_h = G_{\la_h}(\wh C, \wh{C}_{J_h,\bar{A}_h})$, ~$\beta_h= G_{\la_h}(\wh{C}_{U_h, I}, \wh{C}_{J_h,\bar{A}_h})$, ~$\eta_h = G_{\la_h}(\wh C, \wh{C}_{U_h, I})$. When 
 $$\la_0 = \frac{\kappa^2}{\min(t,1)}, \quad q_1 \geq\frac{5 \kappa^2 q_2}{q(1+t)}, \quad q_2 \geq 12 q {(2t + 1)^2 \over t^2}  (1+t)\log \frac{12 H n}{\delta},$$ 
 then the following holds with probability $1-\delta$: for any $h \in \{0, \dots, H\}$ 
 \eqal{\label{eq:def-event-Eh}
 	\begin{split}
 		a)&\qquad \frac{1}{T} \emplev(x, \la_h) ~\leq~ \emplev_{J_h, \bar{A}_h}(x) ~\leq~ \min(T,2)\emplev(x, \la_h), ~~~ \forall x \in X, \\
 		b)&\qquad d_h  ~~\leq~~ 3q~\deff(\la_h) ~\vee ~ 10q, ~~\textrm{and}~~ |J_h| \leq q_2(3q\deff(\la_h) \vee 10q). \\
 		c)&\qquad \beta_h \leq  \frac{7}{11c_T}, ~~\eta_h \leq \frac{3}{11c_T},~~\nu_h \leq \frac{1}{c_T}.
 	\end{split}
 }
where $T = 1+t$ and $c_T = 2 + 1/(T-1)$.
 \et
 \bpr
 Let $H$, $c_T$, $q$ and $\la_h, U_h, J_h, A_h, d_h, P_h=(p_{h,k})_{k=1}^{R_h}$, for $h \in \{0, \dots, H\}$ as defined in Alg.~\ref{alg:est-ridge-lev-scores} and define $\tau = \delta/(3H)$. Now we are going to define some events and we prove a recurrence relation that they satisfy. Finally we unroll the recurrence relation and bound the resulting events in probability.
  \paragraph{Definitions of the events}
 Now we are going to define some events that will be useful to prove the theorem.
 Denote with $E_h$ the event such that the conditions in Eq.~\eqref{eq:def-event-Eh}-(a) hold for $J_h, A_h, U_h$.
 Denote with $F_h$ the event such that
 $$\frac{n}{R_h} \sum_{u \in U_h} \emplev(x_u, \la_{h-1}) \leq \frac{6}{5}\deff(\la).$$
 Denote with $B_{1, h}$ the event such that 
 $\beta_h$, satisfies 
\eqal{\label{eq:cond-beta1}
\beta_h \leq \sqrt{\frac{4 s_h \log \frac{4n}{\tau}}{M_h}}, \quad \textrm{with} \quad s_h := \sup_{k \in \{1,\dots, R_h\}} \frac{1}{R_h p_{h,k}}\|(\widehat{C}_{U_h, I} + \la_h I)^{-1/2}K_{x_{u_k}}\|^2.
}
 Denote with $B_{2,h}$ the event such that $\eta_h$, satisfies 
 $$\eta_h \leq \sqrt{\frac{4 \kappa^2 n \log \frac{\kappa^2}{\la_h \tau}}{(\la_h n + \kappa^2) R_h}}.$$
 
 \paragraph{First bound for $s_h$.}
 Note that, by definition of $p_{h,k}$, that is, by Prop.~\ref{prop:char-emplev-op}
 $$p_{h,k} = n \alglev_{J_{h-1},A_{h-1}}(x_{u_k}, \la_h)/(d_h R_h) = n \emplev_{J_{h-1},\bar{A}_{h-1}}(x_{u_k}, \la_h)/(d_h R_h),$$
 so 
 $$ s_h = \sup_{k \in \{1,\dots, R_h\}} \frac{d_h \|(\widehat{C}_{U_h, I} + \la_h I)^{-1/2}K_{x_{u_k}}\|^2}{n \emplev_{J_{h-1},\bar{A}_{h-1}}(x_{u_k}, \la_h)} = \sup_{u \in U_h} \frac{d_h \emplev_{U_h, I}(x_u, \la_h)}{\emplev_{J_{h-1},\bar{A}_{h-1}}(x_u, \la_h)},$$
 where the last step consists in apply the definition of $\emplev_{U_h, I}$.
 By applying Lemma~\ref{lm:J1A1toJA} and \ref{lm:emplev-lala1} to $\emplev_{U_h,I}(x, \la_h)$, we have 
 \eqals{
 	\emplev_{U_h,I}(x, \la_h) &\leq \frac{1}{1-\eta_h}\emplev(x, \la_h) \leq \frac{\la_{h-1}}{\la_h(1-\eta_h)} \emplev(x,\la_{h-1})
 }
 and analogously by applying Lemma~\ref{lm:emplev-lala1} to $\emplev_{J_{h-1}, \bar{A}_{h-1}}(x, \la_h)$, we have $\emplev_{J_{h-1}, \bar{A}_{h-1}}(x, \la_h) \geq \emplev_{J_{h-1}, \bar{A}_{h-1}}(x, \la_{h-1})$.
 So, by extending the $\sup$ of $s_h$ to the whole $\X$, we have
 $$s_h \leq d_h \sup_{x \in \X} \frac{\emplev_{U_h, I}(x, \la_h)}{\emplev_{J_{h-1}, \bar{A}_{h-1}}(x, \la_h)} \leq \frac{\la_{h-1}d_h}{\la_h(1-\eta_h)} \sup_{x \in \X} \frac{\emplev(x, \la_{h-1})}{\emplev_{J_{h-1}, \bar{A}_{h-1}}(x, \la_{h-1})}.
 $$
 
 Now we are ready to prove the recurrence relation, for $h \in \{1,\dots H\}$,
 $$E_h \supseteq B_{1,h} \cap B_{2,h} \cap E_{h-1} \cap F_h.$$
 
 \paragraph{Analysis of $E_0$.}
 Note that, since $\|\widehat{C}\| \leq \kappa^2$, then 
 $\frac{1}{\kappa^2 + \la} I \preceq (\widehat{C} + \la I)^{-1} \preceq \frac{1}{\la}$, so for any $x \in X$ the following holds
 $$\frac{K(x,x)}{(\kappa^2 + \la)n} \leq \emplev(x,\la) \leq \frac{K(x,x)}{\la n}.$$
 Since $\la_0 = \frac{\kappa^2}{\min(2,T)-1}$ and  $\emplev_{\emptyset, []}(x,\la_0) = \frac{K(x,x)}{\la_0n}$, we have
 $$\frac{1}{T} \emplev(x, \la_0) \leq \frac{1}{T}\frac{K(x,x)}{\la n} \leq \ell_{\emptyset, []}(x, \la_0) = \frac{K(x,x)}{\la_0 n} = \frac{\min(2,T) K(x,x)}{(\kappa^2 + \la_0) n} \leq \min(2,T) \emplev(x,\la_0).$$
 Setting conventionally $d_0, \nu_0, \eta_0, \beta_0 = 0$ (they are not used by the algorithm or the proof), we have that $E_0$ holds everywhere and so, with probability $1$. 
 
 \paragraph{Analysis of $E_{h-1} \cap B_{1,h} \cap B_{2,h}$.}
 First note that under $E_{h-1}$, the following holds $\emplev_{J_{h-1}, \bar{A}_{h-1}}(x, \la_{h-1}) \geq \frac{1}{T}\emplev(x, \la_{h-1})$ and so
 $$s_h  \leq \frac{\la_{h-1} d_h}{\la_h(1-\eta_h)} \sup_{x \in \X} \frac{\emplev(x, \la_{h-1})}{\emplev_{J_{h-1}, \bar{A}_{h-1}}(x, \la_{h-1})} \leq \frac{\la_{h-1} d_h}{\la_h(1-\eta_h)} \sup_{x \in \X} \frac{\emplev(x, \la_{h-1})}{\frac{1}{T}\emplev(x, \la_{h-1})} \leq \frac{T\la_{h-1} d_h}{\la_h(1-\eta_h)}.$$
 Now note that under $B_{2,h}$, by applying the definition of $R_h$ in Alg.~\ref{alg:est-ridge-lev-scores}, by the condition on $q_1$, we have $$\eta_h \leq \sqrt{\frac{4 \kappa^2 n \log \frac{\kappa^2}{\la_h \tau}}{(\la_h n + \kappa^2) R_h}} \leq \sqrt{\frac{4 \log \frac{\kappa^2}{\la_h \tau}}{q_1}}\leq 3/(11c_T) \leq 3/22.$$
 So under $B_{1,h} \cap B_{2,h} \cap E_{h-1}$ and the fact that $q= \frac{\la_{h-1}}{\la_{h}}$, we have $s_h \leq \frac{T\la_{h-1} d_h}{\la_h(1-\eta_h)} \leq (8/7)qTd_h$ and so, since $M_h = q_2 d_h$, by the condition on $q_2$, we have
 $$\beta_h \leq \sqrt{\frac{4 s_h \log \frac{4n}{\tau}}{M_h}} \leq \sqrt{\frac{(32/7) q T d_h \log \frac{4n}{\tau}}{M_h}} = \sqrt{\frac{(32/7) q T \log \frac{4n}{\tau}}{q_2}} < \frac{7}{11c_T},$$
 where in the last step we used the definition of $M_h$ in Alg.~\ref{alg:est-ridge-lev-scores}. 
 Then, since under $B_{1,h} \cap B_{2,h} \cap E_{h-1}$ we have that $\beta_h \leq 7/(11c_T)$, $\eta_h \leq 3/(11c_T) \leq 3/22$, then, by applying \Cref{prop:split-diff} to $\nu_h$ w.r.t. $\eta_h, \beta_h$, we have 
 $$\nu_h \leq \eta_h + (1+\eta_h)\beta_h\leq \left(\frac{3}{11} + \left(1+\frac{3}{22}\right)\frac{7}{11}\right) \frac{1}{c_T} < \frac{1}{c_T}.$$
 Then $\frac{1}{T}   \leq \frac{1-2\nu_h}{1-\nu_h}$ and $\frac{1}{1-\nu_h} \leq \min(T,2)$, so by applying Thm.~\ref{thm:analytical-dec}, we have
 $$ \frac{1}{T}  \emplev(x, \la_h)   \leq \emplev_{J_h, \bar{A}_h}(x,\la_h) \leq \min(T,2) \emplev(x, \la_h).$$
 
 \paragraph{Analysis of $E_{h-1} \cap F_h$.}
 First note that under $E_{h-1}$ the following holds $\emplev_{J_{h-1}, \bar{A}_{h-1}}(x, \la_{h-1}) \leq \min(T,2)\emplev(x, \la_{h-1})$, so, by applying Lemma~\ref{lm:emplev-lala1} to $\emplev_{J_{h-1}, \bar{A}_{h-1}}(x, \la_h)$, we have
 $$ d_h = \frac{n}{R_h} \sum_{u \in U_h} \emplev_{J_{h-1}, \bar{A}_{h-1}}(x_u, \la_h) \leq \frac{\la_{h-1}n}{\la_h R_h} \sum_{u \in U_h} \emplev_{J_{h-1}, \bar{A}_{h-1}}(x_u, \la_{h-1}) \leq  \frac{2\la_{h-1}n}{\la_h R_h} \sum_{u \in U_h} \emplev(x_u, \la_{h-1}).$$
 Moreover under $F_h$, we have $\frac{n}{R_h} \sum_{u \in U_h} \emplev(x_u, \la_{h-1}) \leq \max(5, \frac{6}{5} \deff(\la_{h-1}))$, so, under $E_{h-1} \cap F_h$, we have 
 $$ d_h \leq 2q \max(5,~(6/5)\deff(\la_{h-1})) \leq \max(10q, 3q\deff(\la_h)).$$
 This implies that
 $$|J_h| = M_h = q_2 d_h \leq q_2\max(10q, 3q\deff(\la_h))$$ 
 \paragraph{Unrolling the recurrence relation.}
 The two results above imply $E_h \supseteq B_{1,h} \cap B_{2,h} \cap E_{h-1} \cap F_h$. Now we unroll the recurrence relation, obtaining
 $$
 E_h \supseteq E_0 \cap (\cap_{j=1}^{h} F_j) \cap (\cap_{j=1}^{h} B_{1,j}) \cap (\cap_{j=1}^{h} B_{2,j}),$$
 so by taking their intersections, we have
 \eqal{\label{eq:intersection-events}
 \cap_{h=0}^H E_h \supseteq  E_0 \cap (\cap_{j=1}^{H} F_j) \cap (\cap_{j=1}^{H} B_{1,j}) \cap (\cap_{j=1}^{H} B_{2,j}).
}
 
 \paragraph{Bounding $B_{1,h}, B_{2,h}, F_{h}$ in high probability}
 Let $h \in  [H]$. The probability of the event $B_{1,h}$ can be written as $\mathbb{P}(B_{1,h}) = \int \mathbb{P}(B_{1,h}| U_h, P_h) d\mathbb{P}(U_h, P_h)$. Now note that $\mathbb{P}(B_{1,h}| U_h, P_h)$ is controlled by Lemma~\ref{lm:concentrate-beta1}, that proves that for any $U_h, P_h$, the probability of $\mathbb{P}(B_{1,h}| U_h, P_h)$ is at least $1-\tau$. Then $$\mathbb{P}(B_{1,h}) = \int \mathbb{P}(B_{1,h}| U_h, P_h) d\mathbb{P}(U_h, P_h) \geq \inf_{U_h} \mathbb{P}(B_{1,h}| U_h, P_h) \geq 1- \tau.$$
 To see that $\mathbb{P}(B_{1,h}| U_h, P_h)$ is controlled by \Cref{lm:concentrate-beta1}, note that, since $|U_h|$ is exactly $R_h$, by definition of $\bar{A}_h$ and $A_h$
 $$\bar{A}_h = \frac{|J_h|}{n} A_h = |U_h|~ \textrm{diag}(p_{j_1},\dots, p_{j_{|J_h|}}),$$
 that is exactly the condition on the weights required by Lemma~\ref{lm:concentrate-beta1} which controls exactly \Cref{eq:cond-beta1}.
Finally $B_{2,h},F_h$ are directly controlled respectively by \Cref{lm:concentrate-beta2,lm:concentrate-Fh} and so hold with probability at least $1-\tau$ each. Finally note that $E_0$ holds with probability $1$. So by taking the intersection bound according to \Cref{eq:intersection-events}, we have that $\cap_{h=0}^H E_h$ holds at least with probability $1-3H\tau$.
 \epr
 
 \subsection{Proof for \Cref{alg:mahoneyplus-nystrom-fast}}
 
 \blm\label{lemma:controlling-V}
 Let $\la > 0$, $n \in \N$, $\delta \in (0,1]$. Let $(x_i)_{i=1}^n \subseteq \X$. Let $b \in (0,1]$ and $p_1,\dots, p_n \in (0,b]$. Let $u_1,\dots u_n$ sampled independently and uniformly on $[0,1]$. Let $v_j$ be independent $Bernoulli(p_j/b)$ random variables, with $j \in [n]$. Denote by $z_j$ the random variable $z_j = 1_{u_j \leq b} v_j$. Finally, let the random set $J$ containing $j$ iff $z_j = 1$. Let $A = \frac{n}{|J|}(p_{j_1},\dots,p_{j_{|J|}})$, where $j_1,\dots, j_{|J|}$ are the sorting of $J$. Then the following holds with probability at least $1-\delta$
 $$G_\la(\wh{C}, \wh{C}_{J,A}) \leq \frac{2 s \eta}{3n} + \sqrt{\frac{2 s \eta}{n}}, \quad \textrm{with}\quad s = \sup_{i \in [n]} \frac{1}{p_{i}}\|(\wh{C} + \la I)^{-1/2} K_{x_{i}}\|_\hh^2,$$
 with $s = \log \frac{4 n}{\delta}$.
 \elm
 \bpr
 Let $\zeta_{i}$ be defined as
 $$\zeta_{i} = \frac{z_{i}}{p_{i}}\frac{1}{n} (\wh{C} + \la I)^{-1/2}(K_{x_{i}}\otimes K_{x_{i}})(\wh{C} + \la I)^{-1/2},$$
 for $i \in [n]$, where $z_{i}$ are the Bernoulli random variables computed by \Cref{alg:mahoneyplus-nystrom-fast}.
 First note that 
 \eqals{
 	(\wh{C} + \la I)^{-1/2}\wh{C}_{J,A}(\wh{C} + \la I)^{-1/2} &= \frac{1}{|J|} \sum_{j \in J} \frac{|J|}{n p_{j}} (\wh{C} + \la I)^{-1/2}(K_{x_{i}}\otimes K_{x_{i}})(\wh{C} + \la I)^{-1/2}\\
 	& = \frac{1}{n} \sum_{j \in J} \frac{1}{p_{j}} (\wh{C} + \la I)^{-1/2}(K_{x_{i}}\otimes K_{x_{i}})(\wh{C} + \la I)^{-1/2} \\
 	& = \frac{1}{n} \sum_{i=1}^n \frac{z_i}{p_{j}} (\wh{C} + \la I)^{-1/2}(K_{x_{i}}\otimes K_{x_{i}})(\wh{C} + \la I)^{-1/2} \\
 	& = \sum_{i=1}^n \zeta_i.
 }
 In particular we study the expectation and the variance of $\zeta_i$ to bound $G_\la(\wh{C}, \wh{C}_{J,A})$. By noting that the expectation of $z_i$ is $\expect{z_i} = \expect{1_{u_i \geq b} v_i} = \expect{1_{u_i \geq b}} \expect{v_i} = b \times{\frac{p_i}{b}} = p_i$, for any $i \in [n]$, then
 \begin{align*}
 \expectedvalue ~ \sum_{i=1}^n\zeta_{i} &= \sum_{i=1}^{n} \frac{\expect{z_{i}}}{p_{i}}\frac{1}{n} (\wh{C} + \la I)^{-1/2}(K_{x_{i}}\otimes K_{x_{i}})(\wh{C} + \la I)^{-1/2}\\
 &= \sum_{i=1}^{n}\frac{1}{n} (\wh{C} + \la I)^{-1/2}(K_{x_{i}}\otimes K_{x_{i}})(\wh{C} + \la I)^{-1/2} \\
 & = (\wh{C} + \la I)^{-1/2}\wh{C}(\wh{C} + \la I)^{-1/2} =: W,
 \end{align*}
 Now we will bound almost everywhere $\|\zeta_i\|$ as 
 \begin{align*}
 \|\zeta_{i}\|
 &\leq \sup_{i \in [n]} \frac{z_{i}}{p_{i}}\frac{1}{n}\|(\wh{C} + \la I)^{-1/2} K_{x_{i}}\|_\hh^2
 \leq \frac{1}{n} \sup_{i \in [n]} \frac{1}{p_{i}}\|(\wh{C} + \la I)^{-1/2} K_{x_{i}}\|_\hh^2.
 \end{align*}
 We are ready to apply non-commutative Bernstein inequality (Prop.~\ref{prop:concentration-ClaC} is specific version for this setting), obtaining, with probability at least $1-\delta$
 $$G_\la(\wh{C},\wh{C}_{J,A}) = \|\frac{1}{n} \sum_{i=1}^n (\zeta_i - \expect{\zeta_i})\| \leq \frac{2 s \eta}{3n} + \sqrt{\frac{2 s \eta}{n}},$$
 with $\eta = \log\frac{4 \tr(W)}{\|W\| \delta}$. Finally note that since $\tr(W)/\|W\| \leq \textrm{rank}(W) \leq n$, we have $\eta \leq \log\frac{4 n}{\delta}$.
 \epr

 \blm\label{lemma:controlling-Z}
 Let $\la > 0$, $n \in \N$, $\delta \in (0,1]$. Let $(x_i)_{i=1}^n \subseteq \X$. Let $b \in (0,1]$ and $p_1,\dots, p_n \in (0,b]$. Let $u_1,\dots u_n$ sampled independently and uniformly on $[0,1]$. Let $v_j$ be independent $Bernoulli(p_j/b)$ random variables, with $j \in [n]$. Denote by $z_j$ the random variable $z_j = 1_{u_j \leq b} v_j$. Finally, let the random set $J$ containing $j$ iff $z_j = 1$. Then the following holds with probability at least $1-\delta$
 $$|J| \leq \sum_{i \in [n]}  p_i + (1 + \sqrt{\sum_{i \in [n]}  p_i}) \log \frac{3}{\delta}.$$ 
 \elm
 \bpr
 By definition of $J_h$, note that 
 $$|J| = \sum_{i \in [n]} z_i.$$
 We are going to concentrate the sum of random variables via Bernstein. Any $z_i$ is bounded, by construction, by $1$. Moreover 
 $$\expect{z_i} = \expect{1_{u_i \geq b} v_i} = \expect{1_{u_i \geq b}} \expect{v_i} = b \times{\frac{p_i}{b}} = p_i.$$
 Analogously $\expect{z_i^2} - \expect{z_i}^2 = p_i - p_i^2 \leq p_i$.
 By applying Bernstein inequality, we have
 $$|\sum_{i \in [n]} (z_i - p_i)| \leq \log\frac{2}{\delta} + \sqrt{\log\frac{2}{\delta}\sum_{i \in [n]}  p_i},$$
 with probability $1-\delta$. Then with the same probability,
 $$|J| \leq \sum_{i \in [n]}  p_i + (1 + \sqrt{\sum_{i \in [n]}  p_i}) \log \frac{3}{\delta}.$$ 
 \epr

  \bt\label{thm:alg-mplus-extended-form}
 Let $n \in \N$, $(x_i)_{i=1}^n \subseteq X$. Let $\delta \in (0,1]$, ~$t,q > 1$, $\la > 0$ and $H, d_h, \la_h, J_h, A_h$ as in Alg.~\ref{alg:mahoneyplus-nystrom-fast}. Let $\nu_h = G_\la(\widehat{C}, \widehat{C}_{J_h,\bar{A}_h})$. 
 When 
 $$\la_0 = \frac{\kappa^2}{\min(t,1)}, \quad q_1 \geq 2Tq(1+ 2/t) \log \frac{4n }{\delta}$$
 then, the following holds with probability $1-\delta$: for any $h \in \{0, \dots, H\}$ 
 \eqal{\label{eq:def-event-Eh-mplus}
 	\begin{split}
 		a)&\qquad \frac{1}{T} \emplev(x, \la_h) ~\leq~ \emplev_{J_h, \bar{A}_h}(x) ~\leq~ \min(T,2)\emplev(x, \la_h), ~~~ \forall x \in X, \\
 		b)&\qquad |J_h|  ~~\leq~~ 3 q_1 \min(T,2) \left(5 \vee \deff(\la_h)\right) \log\frac{6H}{\delta},  \\
 		c)&\qquad \nu_h \leq \frac{1}{c_T}.
 	\end{split}
 }
 where $T = 1+t$ and $c_T = 2 + 1/(T-1)$.
 \et
 \bpr
 Let $H$, $c_T$, $q$ and $\la_h, J_h, A_h$, $(p_{h,i})_{i=1}^n$ for $h \in \{0, \dots, H\}$ as defined in Alg.~\ref{alg:mahoneyplus-nystrom-fast} and define $\tau = \delta/(2H)$. Now we are going to define some events and we prove a recurrence relation that they satisfy. Finally we unroll the recurrence relation and bound the resulting events in probability.
  \paragraph{Definitions of the events}
 Now we are going to define some events that will be useful to prove the theorem.
 Denote with $E_h$ the event such that the conditions in Eq.~\eqref{eq:def-event-Eh-mplus}-(a) hold for $J_h, \bar{A}_h$.
 Denote with $Z_h$ the event such that
 $$|J_h| \leq  \sum_{i \in [n]}  p_{h,i} + (1 + (\sum_{i \in [n]}  p_{h,i})^{1/2}) \log \frac{3}{\tau}.$$
 Denote with $V_{h}$ the event such that 
 $\nu_h := G_{\la_h}(\widehat{C}_{U,I}, \widehat{C}_{J_h, A_h})$, satisfies 
 \eqal{\label{eq:cond-nu-mplus}
 	\nu_h \leq s_h \log \frac{8\kappa^2}{\la_h \tau} + \sqrt{2 s_h \log \frac{8\kappa^2}{\la_h \tau}}, \quad \textrm{with}\quad s_h = \sup_{i \in [n]} \frac{1}{n p_{h,i}}\|(\wh{C} + \la_h I)^{-1/2} K_{x_{i}}\|_\hh^2.
 }
 \paragraph{Analysis of $s_h$.}
 Note that, by definition of $p_{h,i}$, for \Cref{alg:mahoneyplus-nystrom-fast}, and of $\emplev$, we have
 so 
 $$ s_h = \sup_{i \in [n]} \frac{1}{n p_{h,i}}\|(\wh{C} + \la_h I)^{-1/2} K_{x_{i}}\|_\hh^2 =\sup_{i \in [n]} \frac{\emplev(x_i,\la_i)}{q_1\alglev_{J_h,A_h}(x_i)} =\sup_{i \in [n]} \frac{\emplev(x_i,\la_i)}{q_1 \emplev_{J_h,\bar{A}_h}(x_i)}.$$
 with $\bar{A}_h = \frac{n}{|J|} A_h$, where the last step is due to the equivalence between $\alglev$ and $\emplev$ in \Cref{prop:char-emplev-op}.
 
 Now we are ready to prove the recurrence relation, for $h \in \{1,\dots H\}$,
 $$E_h \supseteq V_h \cap Z_{h} \cap E_{h-1}.$$
 
 \paragraph{Analysis of $E_0$.}
 Note that, since $\|\widehat{C}\| \leq \kappa^2$, then 
 $\frac{1}{\kappa^2 + \la} I \preceq (\widehat{C} + \la I)^{-1} \preceq \frac{1}{\la}$, so for any $x \in X$ the following holds
 $$\frac{K(x,x)}{(\kappa^2 + \la)n} \leq \emplev(x,\la) \leq \frac{K(x,x)}{\la n}.$$
 Since $\la_0 = \frac{\kappa^2}{\min(2,T)-1}$ and  $\emplev_{\emptyset, []}(x,\la_0) = \frac{K(x,x)}{\la_0n}$, we have
 $$\frac{1}{T} \emplev(x, \la_0) \leq \frac{1}{T}\frac{K(x,x)}{\la n} \leq \ell_{\emptyset, []}(x, \la_0) = \frac{K(x,x)}{\la_0 n} = \frac{\min(2,T) K(x,x)}{(\kappa^2 + \la_0) n} \leq \min(2,T) \emplev(x,\la_0).$$
 Setting conventionally $d_0, \nu_0, \eta_0, \beta_0 = 0$ (they are not used by the algorithm or the proof), we have that $E_0$ holds everywhere and so, with probability $1$. 
 
 \paragraph{Analysis of $E_{h-1} \cap V_{h}$.}
 Note that under $E_{h-1}$, we have $\emplev_{J_{h-1}, \bar{A}_{h-1}}(x, \la_{h-1}) \geq \frac{1}{T}\emplev(x, \la_{h-1})$, so
 \eqals{
 s_h &= \sup_{i \in [n]} \frac{\emplev(x_i,\la_h)}{q_1\emplev_{J_h,\bar{A}_h}(x_i,\la_{h-1})} \leq T\sup_{i \in [n]} \frac{\emplev(x_i,\la_h)}{q_1\emplev(x_i, \la_{h-1})} \\
 &\leq \frac{T \la_{h-1}}{\la_{h}} \sup_{i \in [n]} \frac{\emplev(x_i,\la_{h-1})}{q_1\emplev(x_i, \la_{h-1})} = \frac{T \la_h}{q_1\la_{h-1}} = \frac{Tq}{q_1},
 }
 where we used the fact that $\emplev(x_i, \la_{h}) \leq \frac{\la_{h-1}}{\la_{h}} \emplev(x_i, \la_{h-1})$, via \Cref{lm:emplev-lala1}.
 In particular since we are in $V_h$, this means that, since $q_1 \geq 2Tq(1+ 2/t) \log \frac{4n }{\delta}$, we have
 \eqal{
 	\nu_h &\leq \frac{Tq}{q_1} \log \frac{8\kappa^2}{\la_h \tau} + \sqrt{2 \frac{Tq}{q_1} \log \frac{8\kappa^2}{\la_h \tau}} \leq (4+2t^{-1})^{-2} + \sqrt{2/(4+2t^{-1})^2} \\
 &\leq (1/8 + \sqrt{1/8})(2+t^{-1})^{-1} \leq  \frac{1}{2c_T}.
}
Then $\frac{1}{T} \leq \frac{1-2\nu_h}{1-\nu_h}$ and $\frac{1}{1-\nu_h} \leq \min(T,2)$, so by applying Thm.~\ref{thm:analytical-dec}, we have
$$ \frac{1}{T}  \emplev(x, \la_h)   \leq \emplev_{J_h, \bar{A}_h}(x,\la_h) \leq \min(T,2) \emplev(x, \la_h).$$

 \paragraph{Analysis of $E_{h-1} \cap Z_h$.}
 First consider $\sum_{i \in [n]} p_{h,i}$. By the fact that $\alglev_{J_{h-1},A_{h-1}} = \emplev_{J_{h-1},\bar{A}_{h-1}}$, by \Cref{prop:char-emplev-op}, we have
\eqals{
	\sum_{i \in [n]} p_{h,i} &= q_1 \sum_{i \in [n]} \alglev_{J_{h-1},A_{h-1}}(x_i, \la_{h}) = q_1 \sum_{i \in [n]} \emplev_{J_{h-1},\bar{A}_{h-1}}(x_i, \la_{h}) \\
	& \leq q_1 \frac{\la_{h-1}}{\la_{h}}\sum_{i \in [n]} \emplev_{J_{h-1}, \bar{A}_{h-1}}(x_i, \la_{h-1}),  \leq q_1 \min(T,2) \frac{\la_{h-1}}{\la_{h}}\sum_{i \in [n]} \emplev(x_i, \la_{h-1}), \\
	&  \leq q_1 \min(T,2) \frac{\la_{h-1}}{\la_{h}}\sum_{i \in [n]} \emplev(x_i, \la_h) = q_1 \min(T,2) \deff(\la_h),
}
where we applied in order (1) \Cref{lm:emplev-lala1}, to bound $\emplev_{J_{h-1},\bar{A}_{h-1}}(x_i, \la_{h})$ in terms of $\emplev_{J_{h-1}, \bar{A}_{h-1}}(x_i, \la_{h-1})$, (2) the fact that we are in the event $E_{h-1}$ and so $\emplev_{J_{h-1}, \bar{A}_{h-1}}(x_i, \la_{h-1}) \leq \min(T,2) \emplev(x_i, \la_{h-1})$, then (3) again \Cref{lm:emplev-lala1} to bound $\emplev(x_i, \la_{h-1})$ w.r.t. $\emplev(x_i, \la_{h})$, and (4) finally the definition of $\deff(\la_h)$.

Now if $\deff(\la_h) \leq 10$, we have that
$$ \sum_{i \in [n]}  p_{h,i} + (1 + (\sum_{i \in [n]}  p_{h,i})^{1/2}) \log \frac{3}{\tau} \leq 15 q_1 \min(T,2) \log \frac{3}{\tau}.$$
If $\deff(\la_h) > 10$, we have that
$$\sum_{i \in [n]}  p_{h,i} + (1 + (\sum_{i \in [n]}  p_{h,i})^{1/2}) \log \frac{3}{\tau} \leq 3 \deff(\la_h) q_1 \min(T,2) \log \frac{3}{\tau}.$$
So under $E_{h-1} \cap Z_h$, we have that
$$|J| \leq 3 q_1 \min(T,2) \left(5 \vee \deff(\la_h)\right) \log \frac{3}{\tau}.$$

 \paragraph{Unrolling the recurrence relation.}
 The two results above imply $E_h \supseteq V_{h} \cap Z_{h} \cap E_{h-1}$. Now we unroll the recurrence relation, obtaining
 $$
 E_h \supseteq E_0 \cap (\cap_{j=1}^{h} Z_j) \cap (\cap_{j=1}^{h} V_{j}),$$
 so by taking their intersections, we have
 \eqal{\label{eq:intersection-events-mplus}
 	\cap_{h=0}^H E_h \supseteq  E_0 \cap (\cap_{j=1}^{H} Z_j) \cap (\cap_{j=1}^{H} V_{j}) .
 }
 
 \paragraph{Bounding $V_{h}, Z_{h}$ in high probability}
 Let $h \in  [H]$. Denote by $P_h = (p_{h,j})_{j \in [n]}$. The probability of the event $Z_h$ can be written as $\mathbb{P}(Z_{h}) = \int \mathbb{P}(Z_{h}| P_h) d\mathbb{P}(P_h)$. Now note that $\mathbb{P}(Z_{h}| P_h)$ is controlled by \Cref{lemma:controlling-Z}, that proves that the probability of $\mathbb{P}(Z_{h}| P_h)$ is at least $1-\tau$. Then $$\mathbb{P}(Z_{h}) = \int \mathbb{P}(Z_{h}| P_h) d\mathbb{P}(P_h) \geq \inf_{P_h} \mathbb{P}(Z_{h}|  P_h) \geq 1- \tau.$$
 The probability event $V_h$ is lower bounded by $1-\tau$, via the same reasoning, using \Cref{lemma:controlling-V}.
 Finally note that $E_0$ holds with probability $1$. So by taking the intersection bound according to \Cref{eq:intersection-events-mplus}, we have that $\cap_{h=0}^H E_h$ holds at least with probability $1-3H\tau$.
 \epr

 \subsection{Proof of \Cref{thm:main-appr-lev-scores}}

 \begin{proof}
 	The proof of this theorem splits in the proof for \Cref{alg:est-ridge-lev-scores} that corresponds to \Cref{thm:alg-appr-lev-scores-extended-form} and the proof for \Cref{alg:mahoneyplus-nystrom-fast}, that corresponds to \Cref{thm:alg-mplus-extended-form}. In particular, the result abou leverage scores is expressed in terms of out-of-sample-leverage-scores $\emplev_{J_h, A_h}$ (\Cref{def:out-of-sample-lev-scores-true}).  The desired result, about $\alglev_{J_h, A_h}$, is obtained via \Cref{prop:char-emplev-op}.
 	
 	Note that the two theorems provides stronger guarantees than the ones required by this theorem. We will use only points (a) and (b) of their statements. Moreover they prove the result for the out-of-sample-leverage-scores (\Cref{def:out-of-sample-lev-scores-true}) and here we specify the result only for $x = x_i$, with $i \in [n]$.
 \end{proof}  \section{Theoretical Analysis for Falkon with BLESS}
 
 In this section the FALKON algorithm is recalled in detail. Then it is proved in Thm.~\ref{thm:falkon-to-ny-lsg} that the excess risk of FALKON-BLESS is bounded by the one of \Nystrom{}-KRR. In Thm.~\ref{thm:nystrom-lsg} the learning rates for \Nystrom{}-KRR with BLESS are provided. In Thm.~\ref{thm:generalization-FALKON-LSG} a more general version of Thm.~\ref{thm:FALKON-basic-rates} is provided, taking into account more refined regularity conditions on the learning problem. Finally the proof of Thm.~\ref{thm:FALKON-basic-rates} is derived as a corollary.
 
 \subsection{Definition of the algorithm}
 
 \bd[Generalized Preconditioner]\label{def:generalized-prec}
 Given $\la > 0$, $(\widetilde{x}_j)_{j=1}^M \subseteq X$, $M \in \N$ and $A \in \R^{M\times M}$ positive diagonal matrix,
 we say that $B$ is a {\em generalized preconditioner}, if
 $$B = \frac{1}{\sqrt{n}} A^{-1/2} Q T^{-1} R^{-1},$$
 where $Q \in \R^{M\times q}$ partial isometry with $Q^\top Q = I$ and $q \leq M$, where $T, R \in \R^{q\times q}$ are invertible triangular, and $Q, T, R$ satisfy
 $$A^{-1/2}\Km A^{-1/2} = QT^\top T Q^\top, \quad R = \frac{1}{M} T T^\top + \la I,$$
 with $\Km \in \R^{M \times M}$ defined as $(\Km)_{ij} = K(\widetilde{x}_i, \widetilde{x}_j)$.
 \ed
 
 \bex[Examples of Preconditioners]
 The following are some ways to compute preconditioners satisfying Def.~\ref{def:generalized-prec}
 \begin{enumerate}
 	\item If $\Km$ in the definition above is full rank, then we can choose
 	$$Q = I, \quad T = \textrm{chol}(A^{-1/2}\Km A^{-1/2}), \quad R = \textrm{chol}(\frac{1}{M}TT^\top + \la I),$$
 	where $\textrm{chol}$ is the Cholesky decomposition.
 	\item If $\Km$ is rank deficient, let $q = \textrm{rank}(\Km)$, then
 	$$
 	(Q, Z) = \textrm{qr}(A^{-1/2}\Km A^{-1/2}), \quad T = \textrm{chol}(Q^\top A^{-1/2} \Km A^{-1/2} Q), \quad R = \textrm{chol}(\frac{1}{M}TT^\top + \la I),
 	$$
 	where $\textrm{qr}$ is the QR rank-revealing decomposition. 
 	\item If instead of $\textrm{qr}$ we want to use the eigendecomposition, then let $(\la_j, u_j)_{j=1}^M$ be the eigenvalue decomposition of $A^{1/2} \Km A^{1/2}$ with $\la_1 \geq \dots \geq \la_M \geq 0$ and let $q = \textrm{rank}(\Km)$. Then
 	$$Q = (u_1,\dots, u_q), ~~ T = \textrm{diag}(\sqrt{\la_1},\dots,\sqrt{\la_q}), ~~ R = \textrm{diag}\left(\sqrt{\frac{\la_1}{M}+ \la}, \dots, \sqrt{\frac{\la_q}{M} + \la}\right).$$ 
 \end{enumerate}
 \eex
 
 \bd[Generalized Falkon Algorithm]\label{def:generalized-falkon} Let $\la > 0$ and $t, n, M \in \N$. Let $(x_i, y_i)_{i=1}^n \subseteq X\times Y$ be the dataset. Given $J \subseteq [n]$ let $\widetilde{X}_J = \cup_{j \in J} x_j$ be the selected \Nystrom{} centers and denote by $\{\tilde{x}_1,\dots,\tilde{x}_{|J|}\}$ the points in $\widetilde{X}_J $. Let $A \in \R^{|J|\times|J|}$ be a positive diagonal matrix of weights and $K$ the kernel function. Let $B, q$ be as in Def.~\ref{def:generalized-prec} based on $\widetilde{X}_M$ and $A $. The {\em Generalized Falkon estimator} is defined as follows
 $$\widehat{f}_{\la, J, A, t} = \sum_{i=1}^{|J|} \alpha_i K(x, \widetilde{x}_i), \quad \textrm{with} \quad \alpha = B \beta_t,$$
 where $\beta_t \in \R^q$ denotes the vector resulting from $t$ iterations of the conjugate gradient algorithm applied to the following linear system
 $$W \beta = b, \quad W = B^\top(\Knm^\top \Knm + \la n \Km) B, \quad b= B^\top \Knm^\top y,$$
 with $\Knm \in \R^{n \times M}$, $(\Knm)_{ij} = K(x_i, \widetilde{x}_j)$, and $\Km \in \R^{M\times M}$, $(\Km)_{ij} = K(\widetilde{x}_i, \widetilde{x}_j)$, and with $y = (y_1, \dots, y_n) \in \R^n$.
 \ed

 \bd[Standard \Nystrom{} Kernel Ridge Regression]\label{def:nystrom-krr}
 With the same notation as above,
 the standard \Nystrom{} Kernel Ridge Regression estimator is defined as
 $$\widetilde{f}_{\la, J} = \sum_{i=1}^{|J|} \alpha_i K(x, \widetilde{x}_i), \quad \textrm{with} \quad \alpha = (\Knm^\top \Knm + \la n \Km)^\dag y.$$
 \ed

 \subsection{Main results}
 
 Here, Thm.~\ref{thm:falkon-to-ny-lsg} proves the excess risk of FALKON-BLESS is bounded by the one of \Nystrom{}-KRR. In Thm.~\ref{thm:nystrom-lsg} the learning rates for \Nystrom{}-KRR are provided. In Thm.~\ref{thm:generalization-FALKON-LSG} a more general version of Thm.~\ref{thm:FALKON-basic-rates} is provided, taking into account more refined regularity conditions on the learning problem. Finally the proof of Thm.~\ref{thm:FALKON-basic-rates} is derived as a corollary.
 
 Let $Z_n = (x_i, y_i)_{i=1}^n$ be a dataset and $J \subseteq \{1,\dots, n\}$ and $A \in \R^{|J|\times|J|}$ positive diagonal matrix. In the rest of this section we denote by $\widehat{f}_{\la, J, A, t}$ the Falkon estimator as in Def.~\ref{def:generalized-falkon} trained on $Z_n$ and based on the \Nystrom{} centers $\widetilde{X}_{M} = \cup_{j \in J} \{x_j\}$ and weights $A$ with regularization $\la$ and number of iterations $t$. Moreover we denote by $\widehat{f}_{\la, J}$ the standard \Nystrom{} estimator trained on $Z_n$ and based on the \Nystrom{} centers $\widetilde{X}_{M}$. 
 
 The following theorem is obtained by combining Lemma~2,~3 and Thm.~1 of \cite{rudi2017falkon}, with our Prop.~\ref{prop:AlaBla}.
 \bt\label{thm:falkon-to-ny-lsg}
 Let $\la > 0$, $n \geq 3$, $\delta \in (0,1]$, $t_{\max} \in \N$. Let $Z_n = (x_i, y_i)_{i=1}^n$ be an i.i.d. dataset.
 Let $H$ and $(\la_h)_{h=0}^H, (M_h)_{h=0}^H, (J_h)_{h=0}^H, (A_h)_{h=0}^H$ be outputs of Alg.~\ref{alg:est-ridge-lev-scores} runned with parameter $T = 2$. 
 
 The following holds with probability $1-2\delta$: for each $h \in \{0,\dots, H\}$ such that $0 < \la_h \leq \|C\|$, 
 $${\cal R}(\widehat{f}_{\la_h, J_h, A_h, t}) \leq {\cal R}(\widetilde{f}_{\la_h, J_h}) ~+~ 4 \widehat{v} ~ e^{-t} ~\sqrt{1 + \frac{9\kappa^2}{\la_h n} \log\frac{nH t_{\max}}{\delta}}, \quad \forall t \in \{0,\dots,t_{\max}\},$$
 with $\widehat{v} := \frac{1}{n} \sum_{i=1}^n y_i$.
 \et 
 \bpr 
 Let $\tau = \delta/(t_{\max} H)$ and let $h \in \{1,\dots, H\}$.
 By Lemma~2 and Lemma~3 of \cite{rudi2017falkon}, we have that, when $G_\la(\widehat{C}, \widetilde{C}_{J_h,A_h}) < 1$, with their $\widetilde{C}_{J_h,A_h} = \widehat{C}_{J_h, \bar{A}_h}$ and $\bar{A}_h$ defined as in \cref{thm:alg-appr-lev-scores-extended-form}, then the condition number of $W_h$, that is the preconditioned matrix in Def.~\ref{def:generalized-falkon} with $\la = \la_h$, is controlled by
 $$\textrm{cond}(W_h) \leq \frac{1 + G_{\la_h}(\widetilde{C}_{J_h,A_h}, \widehat{C})}{1-G_{\la_h}(\widetilde{C}_{J_h,A_h}, \widehat{C})}.$$
 Now, by Prop.~\ref{prop:AlaBla}, we have
 $$G_{\la_h}(\widetilde{C}_{J_h,A_h}, \widehat{C}) \leq \frac{G_{\la_h}(\widehat{C}, \widetilde{C}_{J_h,A_h})}{1 - G_{\la_h}(\widehat{C}, \widetilde{C}_{J_h,A_h})}.$$
 So, combining the two results above, we have that when $G_{\la_h}(\widehat{C}, \widetilde{C}_{J_h,A_h}) \leq 1/3$
 $$\textrm{cond}(W_h) \leq \frac{1}{1 - 2 ~G_{\la_h}(\widehat{C}, \widetilde{C}_{J_h,A_h})} \leq 3.$$
 Now denote by $E_{h,t}$ the event such that
 $${\cal R}(\widehat{f}_{\la_h, J_h, A_h, t}) ~~\leq~~ {\cal R}(\widetilde{f}_{\la_h, J_h}) ~+~ 4 \widehat{v}^2 ~ e^{-t} ~\sqrt{1 + \frac{9\kappa^2}{\la_h n} \log\frac{n}{\tau}}.$$
 Since $\textrm{cond}(W_h) \leq 3$, we have that $\log \frac{\sqrt{\textrm{cond}(W_h)}+1}{\sqrt{\textrm{cond}(W_h)}+1} \geq 1$ and 
 so can apply Theorem~1 of \cite{rudi2017falkon} with their parameter $\nu = 1$, obtaining that each $E_{h,t}$, with $t \in \{0,\dots, t_{\max}\}$ hold with probability $1-\tau$.
 So by taking the intersection bound, we know that $E_h : = \cap_{t=0}^{t_{\max}} E_{h,t}$ holds with probability $1-t_{\max} \tau$.
 
 Finally denote by $F_H$ the event: $G_{\la_h}(\widehat{C}, \widetilde{C}_{J_h,A_h}) \leq 1/3$ for any $h \in \{0,\dots, H\}$.
 Note that \Cref{thm:alg-appr-lev-scores-extended-form} states that, by running Alg.~\ref{alg:est-ridge-lev-scores} with $T = 2$, the event $F_H$ holds with probability at least $1-\delta$. 
 
 The desired result correspond to the event $\cap_{h=1}^H E_h \cap F_H$ which, by taking the intersection bound, holds with probability at least $1 - \delta - t_{\max} H \tau$.
 \epr

 \subsection{Result for \Nystrom{}-KRR and BLESS}
 
 We introduce here the ideal and empirical operators that we will use in the following to prove the main results of this work and then we prove learning rates for \Nystrom{}-KRR.
 
 In the following denote with $C:\hh \to \hh$ the linear operator 
 $$C =\int K_x \otimes K_x d\rhox(x),$$
 and, given a set of input-output pairs $\{(x_i,y_i)\}_{i=1}^n$ with $(x_i,y_i)\in\X\times\Y$ independently sampled according to $\rho$ on $\X\times\Y$, we define the empirical counterparts of the operators just defined as $\hat{S}:\hh\to\R^n$ s.t. 
 $$f \in\hh \mapsto \frac{1}{\sqrt{n}}(\scal{K_{x_i}}{f}_\hh)_{i=1}^n \in \R^n,$$ 
 with adjoint $\hat{S}^*:\R^n\to\hh$ s.t. 
 $$v = (v_i)_{i=1}^n\in\R^n \mapsto \frac{1}{\sqrt{n}} \sum_{i=1}^n v_i K_{x_i},$$
 
 Now we introduce some assumption that will be satisfied by the conditions on Thm.~\ref{thm:FALKON-basic-rates}.
 
 \ba\label{asm:noise}
 There exists $B, \sigma > 0$ such that the following holds almost everywhere on $\X$
 $$\expect{|y - \expect{y|x}|^p~\mid~x} \leq \frac{p!}{2} B^{p-2} \sigma^2.$$
 \ea

 \ba\label{asm:source}
 There exists $r \in [1/2, 1]$ and $g \in \hh$ such that
 $$\fh = C^{r-1/2} g,$$
 \ea

 \bt[Generalization properties of \Nystrom{}-RR using BLESS]\label{thm:nystrom-lsg}
 Let $\delta \in (0,1]$ and $\la > 0, n \in \N$.Under Asm.~\ref{asm:noise},~\ref{asm:source}, let the \Nystrom{} estimator as in \Cref{def:nystrom-krr} and assume that $(J_h)_{h=1}^H, (A_h)_{h=1}^H, (\la_h)_{h=1}^H$ is obtained via Alg.~\ref{alg:est-ridge-lev-scores} or \ref{alg:mahoneyplus-nystrom-fast}. When $ \frac{9\kappa^2}{n}\log\frac{n}{\delta} \leq \la \leq \|C\|$, then the following holds with probability $1-4\delta$
 $${\cal R}(\widetilde{f}_{\la_h, J_h}) \leq 8\|g\|_\hh \left( \frac{B \log \frac{2}{\delta}}{n \sqrt{\la_h}}  + \sqrt{\frac{\sigma^2 \deff(\la_h)\log \frac{2}{\delta}}{n}} + \la_h^{1/2+v}\right).$$ 
 \et
 \bpr
 The proof consists in following the decomposition in Thm.~1 of \cite{rudi2015less}, valid under Asm.~\ref{asm:source} and using our set $J_h$ to determin the \Nystrom{} centers.
 First note that under \Cref{asm:source}, there exists a function $f_\hh \in \hh$, such that ${\cal E}(f_\hh) = \inf_{f \in \hh} {\cal E}(f)$ (see \cite{caponnetto} and also \cite{steinwart2009optimal,lin2018optimal}).
According to Thm.~2 of \cite{rudi2015less}, under Asm.~\ref{asm:source}, we have that
 $${\cal R}(\widetilde{f}_{\la_h, J_h})^{1/2} \leq  q (\underbrace{{\cal S}(\la_h, n)}_{\textrm{Sample error}} \; + \,{\underbrace{{\cal C}(M_h)^{1/2 + v}}_{\textrm{Computational error}}}\, + \underbrace{\la_h^{1/2+v}}_{\textrm{Approximation error}}),$$
 where ${\cal S}(\la, n) = \nor{(C+\la I)^{-1/2}(\Sn^* \yn - \Cn f_\hh)}$ and ${\cal C}(M_h) = \nor{(I - P_{M_h})(C+\la I)^{1/2}}^2$ with $P_{M_h} = \widehat{C}_{J_h, I} \widehat{C}_{J_h, I}^\dag$. 
 Moreover $q = \|g\|_\hh ( \beta^2 \vee (1 + \theta\beta))$, $\beta = \nor{(\Cn + \la I)^{-1/2} (\C + \la I)^{1/2}}$, $\theta = \nor{(\Cn + \la I)^{1/2}(C + \la I)^{-1/2}}$.
 
 The term ${\cal S}(\la_h, n)$ is controlled under Asm.~\ref{asm:noise} by Lemma~4 of the same paper, obtaining
 $${\cal S}(\la, n) \leq \frac{B \log \frac{2}{\delta}}{n \sqrt{\la_h}} + \sqrt{\frac{\sigma^2 \deff(\la_h)\log \frac{2}{\delta}}{n}},$$
with probability at least $1-\delta$.
The term $\beta$ is controlled by Lemma~5 of the same paper, 
$$\beta \leq 2,$$
with probability $1-\delta$ under the condition on $\la$.
Moreover
 $$\theta^2 = \|(C+\la I)^{-1/2} \widehat{C} (C+\la I)^{-1/2}\| \leq 1 + \|(C+\la I)^{-1/2} (\widehat{C} - C) (C+\la I)^{-1/2}\|,$$
 where the last term is bounded by $1/2$ with probability $1-\delta$ under the same condition on $\la$, via Prop.~8 and the following Remark~1 of the same paper.

 Now we study the term ${\cal C}(M_h)$ that is the one depending on the result of BLESS. First note that, since $\textrm{diag}(A_h) > 0$, then  
 $$P_{M_h} = \widehat{C}_{J_h, I} \widehat{C}_{J_h, I}^\dag = \widehat{C}_{J_h, \bar{A}_h} \widehat{C}_{J_h, \bar{A}_h}^\dag.$$
 By applying Proposition~3 and Proposition~7 of the same paper, the following holds
 $${\cal C}(M_h) \leq \frac{\la_h}{1 - G_{\la_h}(\wh{C}, \wh{C}_{J_h, \bar{A}_h})}, \leq 2 \la_h,$$
 with probability at least $1-\delta$,
 where we applied Thm.~\ref{thm:alg-appr-lev-scores-extended-form}-(c) and Thm.~\ref{thm:alg-mplus-extended-form}-(c), which control exactly $G_{\la_h}(\wh{C}, \wh{C}_{J_h, \bar{A}_h})$ and prove it to be smaller than $1/2$ in high probability.
 
 Finally by taking the intersection bound of the events above, we have
 $${\cal R}(\widetilde{f}_{\la_h, J_h})^{1/2} \leq  4\|g\|_\hh \left( \frac{B \log \frac{2}{\delta}}{n \sqrt{\la_h}}  + \sqrt{\frac{\sigma^2 \deff(\la_h)\log \frac{2}{\delta}}{n}} + 2\la_h^{1/2+v}\right),$$
with probability $1-4\delta$.
\epr

\bt[Generalization properties of learning with FALKON-BLESS]\label{thm:generalization-FALKON-LSG}
 Let $\delta \in (0,1]$ and $\la > 0, n \geq 3$, $t_{\max} \in \N$. Let $Z_n = (x_i, y_i)_{i=1}^n$ be an i.i.d. dataset.
 Let $H$ and $M_H, J_H, A_H$ be outputs of Alg.~\ref{alg:est-ridge-lev-scores} runned with parameter $T = 2$. Let $y \in [-a/2, a/2]$ almost surely, with $a > 0$. Under~\ref{asm:source}, Let $\la > 0$, $n \geq 3$, $\delta \in (0,1]$, when $ \frac{9\kappa^2}{n}\log\frac{n}{\delta} \leq \la \leq \|C\|$, then the following holds with probability $1-6\delta$
 $${\cal R}(\widehat{f}_{\la, J_H, A_H, t}) \leq 4a ~ e^{-t} ~+~ 32\|g\|^2_\hh \left( \frac{a^2 \log^2 \frac{2}{\delta}}{n^2 \la}  + \frac{a \deff(\la)\log \frac{2}{\delta}}{n} + 2\la^{1+2r}\right), \quad \forall t \in \{0,\dots,t_{\max}\},$$
\et
\bpr
The result is obtained by combining Thm.~\ref{thm:falkon-to-ny-lsg}, with Thm.~\ref{thm:nystrom-lsg} and noting that when $y \in [-a/2, a/2]$ almost surely, then it satisfies Asm.~\ref{asm:noise} with $B, \sigma \leq a$.
\epr

\subsection{Proof of Thm.~\ref{thm:FALKON-basic-rates}}
\bpr
The result is a corollary of Thm.~\ref{thm:generalization-FALKON-LSG}, where we assumed only the existence of $f_\hh$. This correspond to assume Asm.~\ref{asm:source}, with $r=1/2$ and $g = f_\hh$ (see \cite{caponnetto}).
\epr
  \section{More details about BLESS and BLESS-R}
{\bf BLESS (Alg.~\ref{alg:est-ridge-lev-scores}).}~ Here we describe our bottom-up algorithm in detail (see \Cref{alg:est-ridge-lev-scores}).
The central element is using a decreasing list of 
$\{\lambda_h\}_{h=1}^h$, from a given $\la_0 \gg \la$ up to $\la$.
The idea is to iteratively construct a LSG set that approximates well the RLS for a given $\la_h$, based on the accurate RLS computed using a LSG set for $\la_{h-1}$.
The crucial observation of the proposed algorithm is that when $\lambda_{h-1} \geq \lambda_{h}$ then
\begin{align*}
\forall i : \ell(i, \lambda_h) \leq \frac{\la_{h}}{\la_{h-1}}\ell(i,\la_{h-1}), && \deff(\lambda_h) \leq \frac{\la_{h}}{\la_{h-1}}\deff(\lambda_{h-1}),
    \end{align*}
(see \Cref{lm:emplev-lala1}, for more details).
 By smoothly decreasing $\lambda_h$, the LSG at step $h$ will only be a $\la_h/\la_{h-1}$ factor worse than our previous estimate, which is
 automatically compensated by a $\la_h/\la_{h-1}$ increase in the size of the LSG.
Therefore, to maintain an accuracy level for the leverage scores approximation as in Eq.~\eqref{eq:cond-appr-lev} and small space complexity, it is sufficient to select a logaritmically spaced list of $\lambda$'s from $\la_0 = \kappa^2$ to $\la$ (see Thm.~\ref{thm:main-appr-lev-scores}), in order to keep $\la_h/\la_{h-1}$ as a small constant. This implies an extra multiplicative computational cost for the whole algorithm of only $\log(\kappa^2/\la)$.

More in detail, we initialize the Algorithm setting $D_0 = (\emptyset,[])$ to the empty LSG.
Afterwards, we begin our main loop where at every step we reduce $\lambda_h$
by a $q$ factor, and then use $D_{h-1}$ to construct a new LSG $D_h$.
Note that at each iteration we construct a set $J_h$ larger than $J_{h-1}$, which requires computing
$\alglev_{D_{h-1}}(i,\la_h)$ for samples that are not in $J_{h-1}$, and therefore not computed at the previous step.
Computing approximate
leverage scores for the whole dataset would be highly inefficient, requiring $\bigotime( n M_h^2 )$ time
which makes it unfeasible for large $n$. Instead, we show that to achieve the desired accuracy it is sufficient to restrict all our operations
to a sufficiently large intermediate subset $U_h$ sampled uniformly from $[n]$.
After computing $\alglev_{D_{h-1}}(i,\la_h)$ only for points in $U_h$, we select $M_h$ points with replacements
according to their RLS to generate $J_h$. With a similar procedure we update the weights
in $A_h$.
We will see in Thm.~\ref{thm:main-appr-lev-scores}, $|U_h| \propto 1/\lambda_h$ 
is sufficient to guarantee that this intermediate step produces a set satisfying \Cref{eq:cond-appr-lev},
and also takes care of  increasing $|U_h|$ to increase accuracy as $\lambda_h$ decreases.
Moreover the algorithm uses a $M_h \propto \sum_{u \in U_h} \alglev_{D_{h-1}}(i,\la_h)$ that we prove in Thm.~\ref{thm:main-appr-lev-scores}, to be in the order of $\deff(\la_h)$.
In the end, we  return either the final LSG $D_H$ to compute approximations
of $\ell(i,\lambda)$, or  any of the intermediate $D_h$ if we are interested
in the RLSs along the regularization path $\{\lambda_h\}_{h=1}^H$.
\\
\\
{\bf BLESS-R (Alg.~\ref{alg:mahoneyplus-nystrom-fast})}
The second algorithm we propose, is based on the same principles of \Cref{alg:est-ridge-lev-scores}, while simplifying some steps of the procedure. In particular it removes the need to explicitly track the normalization
constant $d_h$ and the intermediate uniform sampling set, by replacing it with \emph{rejection} sampling.
At each iteration $h \in [H]$, instead of drawing the set $U_h$ from a uniform
distribution, and then sampling $J_h$, from $U_h$,
\Cref{alg:mahoneyplus-nystrom-fast} performs a single round of rejection sampling for each column according to the following identity
$$\probability(z_{h,i} = 1) = \probability(z_{h,i} = 1| u_{h,i} \leq \beta_h)\probability(u_{h,i} \leq \beta_h)
= \beta_hp_{h,i}/\beta_h
= p_{h,i} \propto
\alglev_{D_{h-1}}(x_{i}, \la_{h-1}),$$
where $z_{h,i}$ is the r.v. which is $1$ if $i \in [n]$, while $u_{h,i}$ is the probability that the column $i$ passed the rejection sampling step, while $\beta_h$ a suitable treshold which mimik the effect of the set $U_h$.
\\
\\
{\bf Space and time complexity.}~
Note that at each iteration constructing the generator $\alglev_{D_{h-1}}$, requires computing the inverse
$(K_{J_h} + \lambda_h n I)^{-1}$, with $M_h^3$ time complexity, while each of the $R_h$ evaluations $\alglev_{D_{h-1}}(i,\lambda_h)$ takes only $M_h^2$ time. Summing over the $H$ iterations Alg.~\ref{alg:est-ridge-lev-scores} runs in
$\bigotime(\sum_{h=1}^H M_h^3 + R_hM_h^2)$ time. 
Noting that $R_h \simeq 1/\lambda_h$, that $M_h \simeq d_h \leq 1/\lambda_h$,
and that $\sum_h \la_h^{-1} = \sum_h q^{h-H} \la^{-1} = \frac{q- q^{-H}}{q-1} \la^{-1}$, the final cost is $\bigotime\left(\la^{-1} ~\max_{h} M^2_h\right)$ time, and $\bigotime\left(\max_h M_h^2\right)$ space.
Similarly, Alg.~\ref{alg:mahoneyplus-nystrom-fast}
only evaluates $\alglev_{D_{h-1}}$
for the points that pass the rejection steps
which \whp happens only $\bigotime(n\beta_h) = \bigotime(1/\la)$ times, so we have the same time and space complexity of Alg.~\ref{alg:est-ridge-lev-scores}.

\end{document}